%% file: submission.tex
\documentclass[10pt,twocolumn,letterpaper]{article}

\usepackage{cvpr}              

\input{preamble}

\definecolor{cvprblue}{rgb}{0.21,0.49,0.74}
\usepackage[pagebackref,breaklinks,colorlinks,citecolor=cvprblue]{hyperref}
\usepackage{tikz}
\usepackage{pgfplots}
\usepackage{makecell}
\usepackage{multirow}
\usepackage{lipsum}
\usepackage[accsupp]{axessibility}  

\definecolor{yongzhi}{rgb}{0.9,0.,0.5}

\definecolor{yongliang}{rgb}{0.9,0.5,0.}

\def\paperID{5650} 
\def\confName{CVPR}
\def\confYear{2024}
\renewcommand\footnotemark{}

\title{HiPose: Hierarchical Binary Surface Encoding and Correspondence Pruning \\ for RGB-D 6DoF Object Pose Estimation}

\author{%
  \begin{tabular}{ccccc}
    Yongliang Lin\textsuperscript{1*} &
    Yongzhi Su\textsuperscript{2,3*} &
    Praveen Nathan\textsuperscript{2} &
    Sandeep Inuganti\textsuperscript{2,3} &
    Yan Di\textsuperscript{4,5} \\
    Martin Sundermeyer\textsuperscript{5} &
    Fabian Manhardt\textsuperscript{5} &
    Didier Stricker\textsuperscript{2,3} &
    Jason Rambach\textsuperscript{2,†} &
    Yu Zhang\textsuperscript{1,†}
  \end{tabular}
  \\
  \textsuperscript{1} Zhejiang University 
  \textsuperscript{2} German Research Center for Artificial Intelligence (DFKI) \\
  \textsuperscript{3} RPTU Kaiserslautern-Landau
  \textsuperscript{4} Technische Universit\"{a}t M\"{u}nchen 
  \textsuperscript{5} Google\\
  \tt\small{\{lyl1020; zhangyu80\}@zju.edu.cn},
  \tt\small{jason.rambach@dfki.de}\\
  \thanks{*Contributed equally, †Corresponding authors}
  \thanks{Code: \href{https://github.com/lyltc1/HiPose}{https://github.com/lyltc1/HiPose}}
}
\begin{document}
\maketitle

\input{Sections/0_abstract}

\input{Sections/1_intro}
\input{Sections/2_related}
\input{Sections/3_method}

\input{Sections/4_exp}
\input{Sections/5_conclusion}

\section*{Acknowledgements} 
This work was partially supported by STI 2030-Major Projects 2021ZD0201403, in part by NSFC 62088101 Autonomous Intelligent Unmanned Systems and was partially funded by the EU Horizon Europe Framework Program under grant agreement 101058236 (HumanTech) and the German Ministry of Education and Research (BMBF) under Grant Agreements 01IW20002 (SocialWear) and 01IW20009(RACKET).

\clearpage 
{
    \small
    \bibliographystyle{ieeenat_fullname}
    \bibliography{main}
}

\clearpage 
\section{Supplementary Material }
\input{Sections/sup}

\end{document}

%% file: preamble.tex
\usepackage[dvipsnames]{xcolor}


%% file: Sections/0_abstract.tex
\begin{abstract}
In this work, we present a novel dense-correspondence method for 6DoF object pose estimation from a single RGB-D image. 
While many existing data-driven methods achieve impressive performance, they tend to be time-consuming due to their reliance on rendering-based refinement approaches.
To circumvent this limitation, we present HiPose, which establishes 3D-3D correspondences in a coarse-to-fine manner with a hierarchical binary surface encoding. 
Unlike previous dense-correspondence methods, we estimate the correspondence surface by employing point-to-surface matching and iteratively constricting the surface until it becomes a correspondence point while gradually removing outliers.
Extensive experiments on public benchmarks LM-O, YCB-V, and T-Less demonstrate that our method surpasses all refinement-free methods and is even on par with expensive refinement-based approaches. 
Crucially, our approach is computationally efficient and enables real-time critical applications with high accuracy requirements.
\end{abstract}

%% file: Sections/1_intro.tex
\section{Introduction}
\label{sec:intro}

\begin{figure}[t]
\centering
\includegraphics[width=0.475\textwidth]{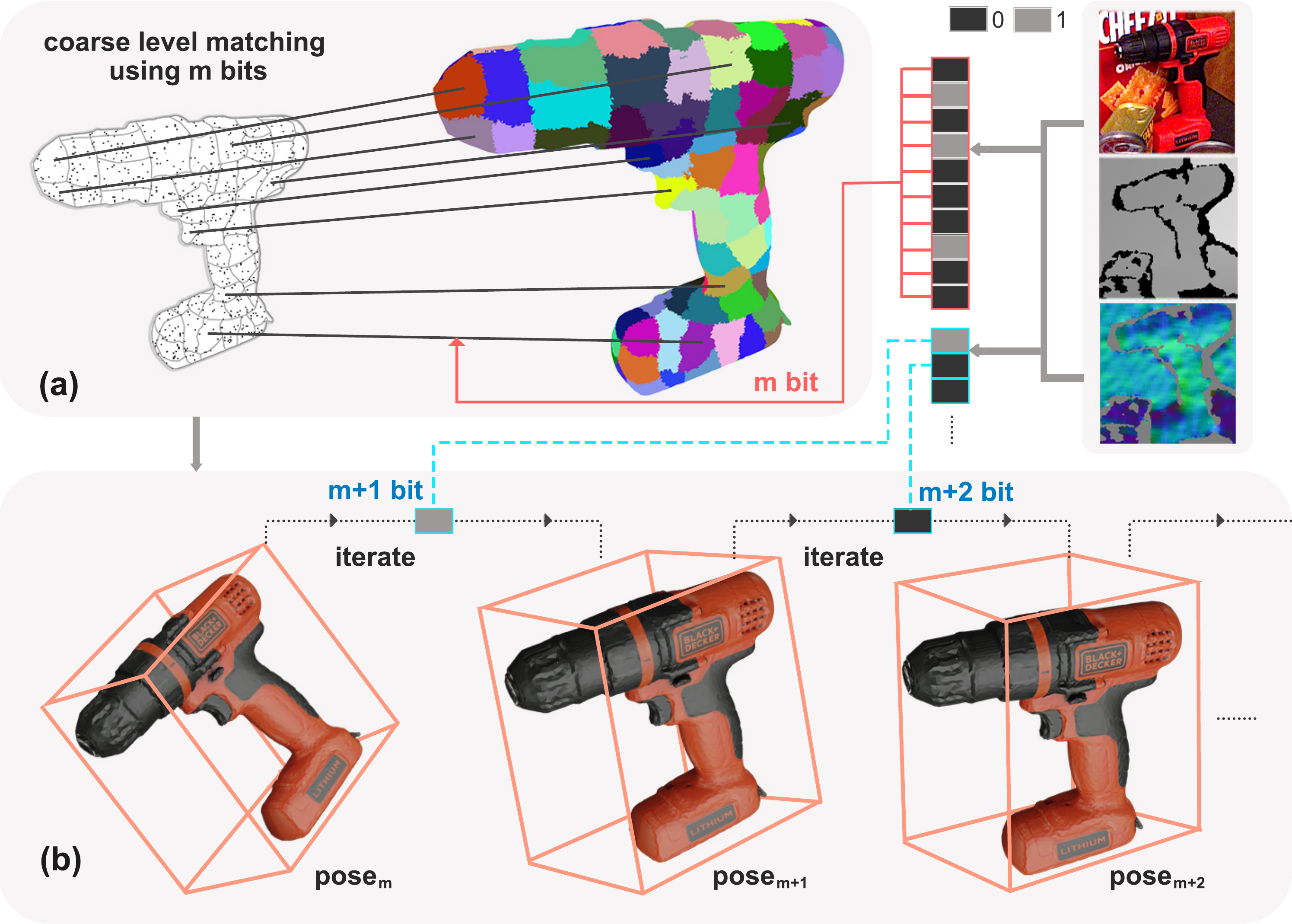}
\caption{\textbf{Illustration of HiPose :} (a) For every point cloud with color and normals as inputs, our network outputs a binary code to establish a correspondence to a sub-surface on the object. (b) With the coarse level matching, we estimate an initial pose $pose_m$. The additional $n$ bits are used for iterative fine-grained matching and pose estimation and gradual outlier rejection. Note that this process is render-free and RANSAC-free, ensuring fast performance of our algorithm.}
\label{fig:teaser}
\end{figure}

Estimating the 6 Degrees of Freedom (6DoF) object pose stands as a fundamental challenge in 3D computer vision. 
This task plays a pivotal role in numerous real-world applications, including augmented reality~\cite{rambach2017poster,su2019deep,zhang2023ddf}, robotic grasping~\cite{zhai2023monograspnet,zhai2023sg}, and autonomous driving~\cite{manhardt2019roi,su2023opa}. 
Despite its significance, achieving accurate 6DoF pose estimation remains challenging, particularly in scenarios characterized by homogeneous object textures and heavy occlusion.

The advent of deep learning has helped to overcome those challenges.
A recent branch of RGB-only works~\cite{zakharov2019dpod,park2019pix2pose,su2022zebrapose} shows promising results for handling occlusion.
Despite these advancements, estimating object pose from RGB images alone remains challenging due to the inherent depth ambiguity in monocular images. 

Analogous to the prediction of 2D-3D correspondence in RGB-only approaches,~\cite{wu2022vote,chen2020g2l} predict sparse 3D-3D correspondences. 
However, most methods with depth input either do not utilize RGB information~\cite{drost2010model,gao2021cloudaae,li2023depth} or rely on RGB images only to segment the object from the background~\cite{konig2020hybrid,gao2021cloudaae,cai2022ove6d,lin2023transpose}, thus discarding valuable RGB features. 
To preserve rich RGB information,~\cite{wang2019densefusion, he2020pvn3d,he2021ffb6d,zhou2023deep} proposed novel feature fusion networks to better leverage RGB and depth information but lag behind in public benchmarks such as BOP~\cite{Sundermeyer2023BOPC2}.
 
In contrast, most current state-of-the-art approaches typically obtain an initial pose using RGB-only methods and then apply a computationally expensive, often iterative, pose refinement step using depth information~\cite{Rusinkiewicz2001EfficientVO,lipson2022coupled}. Directly utilizing RGB-D images to estimate the initial pose promises to yield more precise and reliable object pose estimates. 

In this paper, we aim to fully exploit the detailed information in RGB-D images to estimate accurate object poses without any time-consuming refinement step. Using RGB-D input, we benefit from additional information such as point-to-surface distances. Taking inspiration from the recent work ZebraPose~\cite{su2022zebrapose}, a dense 2D-3D correspondence prediction method, we introduce HiPose, a network that efficiently predicts dense 3D-3D correspondences between the input depth map and the object model. Unlike ZebraPose~\cite{su2022zebrapose}, we process the encoding in a manner that takes better advantage of its coarse-to-fine properties by iteratively removing outliers.
 
Instead of solving the pose using the predicted correspondences within the RANSAC framework, as commonly done in conjunction with the Kabsch algorithm~\cite{umeyama1991least}, we propose a novel and more stable hierarchical correspondence pruning approach, eliminating the need for RANSAC.
Specifically, the coarse-level prediction in the hierarchical binary code output is less error-prone, providing a robust initial pose. This coarse pose helps identifying and removing outlier matches based on point-to-surface distance. Subsequently, we apply a finer-level prediction with each iteration, refining our pose prediction and eliminating outliers at finer levels to enhance accuracy as shown in Figure~\ref{fig:teaser}.

Overall, our contributions can be summarized as follows:
\begin{itemize}
\item We present an approach for estimating object pose that fully exploits RGB-D data, focusing on 3D-3D correspondence matching through hierarchical binary surface encoding.
\item We introduce a RANSAC-free hierarchical correspondence pruning approach for pose estimation through coarse-to-fine sub-surfaces based outlier filtering.
\item Extensive experiments on LM-O, YCB-V and T-LESS datasets demonstrate our method's effectiveness. We achieve state-of-the-art results without any additional refinement, making our approach notably faster than alternative methods and suitable for real-time applications.
\end{itemize}

%% file: Sections/2_related.tex
\section{Related Work}
\label{sec:related}
  We limit our in-depth discussion of related work to the instance-level pose estimation methods based on deep learning where the 3D CAD model of the target object is available during training.

\subsection{RGB-only Pose Estimation}
  Most top-performing RGB object pose estimation methods~\cite{zakharov2019dpod, park2019pix2pose, hodan2020epos, wang2021gdr, di2021so, su2022zebrapose} attempt to establish dense 2D-3D correspondences between 2D coordinates in the RGB image and 3D coordinates on the object surface. The 6D pose is then computed by solving the Perspective-n-Point(PnP) problem~\cite{lepetit2009ep}.
  Dense correspondence-based methods have been shown to outperform the keypoint-based methods~\cite{rothganger20063d, oberweger2018making, peng2019pvnet, zhao2020learning, liu2020keypose} and holistic approaches ~\cite{kehl2017ssd,do2018deep6dpose,xiang2018posecnn,su2021synpo} nowadays, as also demonstrated in the BOP challenge results~\cite{Sundermeyer2023BOPC2}.
  We draw inspiration from ZebraPose~\cite{su2022zebrapose}, a dense correspondence-based method that employs a coarse-to-fine surface encoding to represent correspondences. This approach has demonstrated significant improvements in accuracy, motivating our own idea.
  Overall, RGB-only methods are still limited in performance due to the absence of geometric information.
  
\subsection{Depth-only and RGB-D Pose Estimation}
The development of point cloud processing networks~\cite{qi2017pointnet++, atzmon2018point, li2018pointcnn, hu2020randla} boosted pose estimation methods that exclusively used 3D measurements~\cite{aoki2019pointnetlk, wang2019deep,wang2019prnet, yew2020rpm, yuan2020deepgmr, dang2022learning}. These methods have demonstrated excellent generalization. However, discarding RGB appearance severely limits the performance of these approaches, due to pose ambiguities and exclusion of color features.

RGB-D methods attempt to fuse the information of the RGB and Depth modalities.~\cite{ku2018joint,li2018unified,liang2018deep} treat the depth information as an extra channel of RGB images, which are then fed to a CNN-based network. A more effective utilization of RGB and depth images is to extract features from these two modalities individually, then fuse them for pose estimation~\cite{xu2018pointfusion, qi2018frustum, wang2019densefusion, zhou2020novel, wada2020morefusion, he2020pvn3d, he2021ffb6d, zhou2023deep}. Such approaches benefit from visual information and geometric information and show higher accuracy~\cite{Sundermeyer2023BOPC2}. FFB6D~\cite{he2021ffb6d} designed bidirectional fusion modules to enhance representation of appearance and geometry features. Recently,~\cite{zhou2023deep} proposed a transformer-based fusion network based on FFB6D.

\subsection{Pose Refinement with Depth Information}
An additional pose refinement stage, often in an iterative manner, improves the result significantly. The Iterative Closest Point algorithm (ICP) is typically used as a refinement strategy, making use of depth information to align the estimated object point cloud to the image~\cite{sundermeyer2018implicit,xiang2018posecnn, sundermeyer2020multi}.
PFA~\cite{hu2022perspective} proposes a non-iterative pose refinement strategy by predicting a dense correspondence field between the rendered and real images.
CIR~\cite{lipson2022coupled} iteratively refines both pose and dense correspondence together using a novel differentiable solver layer under a rendering and comparison strategy. However, the rendering is time-consuming.




%% file: Sections/3_method.tex
\section{Method: HiPose}
\label{sec:method}
\begin{figure*}[h]
\centering
\includegraphics[width=\textwidth]{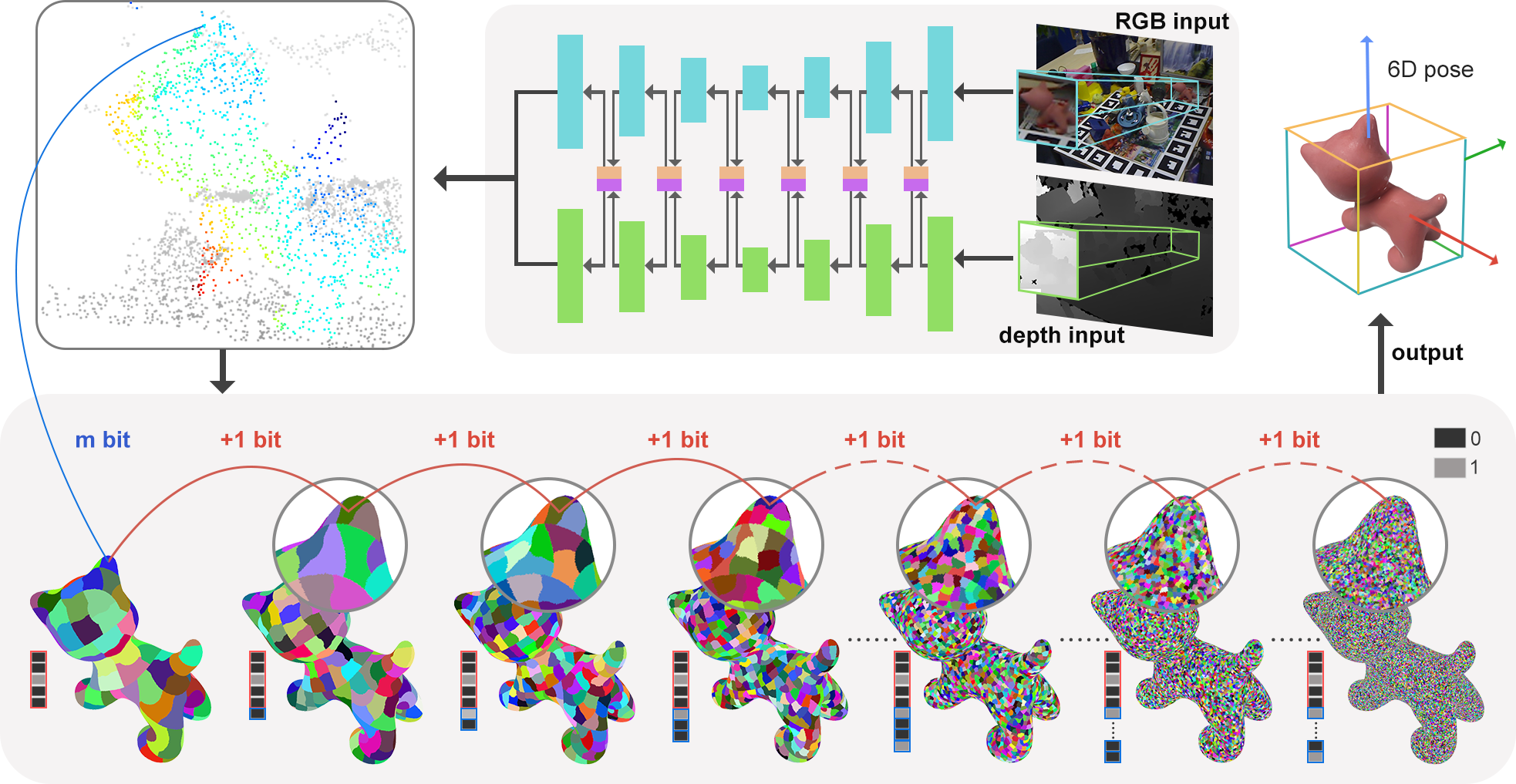}
\caption{\textbf{Overview :} Our framework uses an RGB-D image crop as input and predicts an $m+n$ bits binary code using a full-flow bidirectional fusion network for every point cloud patch on the target object. The first $m$ bit codes point to a relatively coarse surface (blue line), while the final $n$ bit codes are used $n$ times as indicators to perform hierarchical surface partitioning (red lines). Through the iterative process of identifying fine-grained point-to-surface correspondences, the algorithm finally yields an accurately estimated pose. The colored patches on the model represent different surface partitions.}
\label{fig:overview}
\end{figure*}

This section provides a detailed description of the proposed model-based method for 6D object pose estimation. 

Inspired by the successful application of binary codes in the RGB-only setting, we extend the method to encode object surfaces in a coarse-to-fine way in the RGB-D setting.
Our approach consists of a hierarchical binary surface encoding which is fed into a coarse-to-fine pose solver. 
The solver achieves rapid pose estimation through several iterations of surface partitioning and outlier rejection without RANSAC or the need for rendering.

\subsection{Problem Definition and Notation}{\label{sec:notation}}

For each target object in an RGB-D image (RGB image + 2D depth map), our goal is to estimate the transformation between the predefined object coordinate system and the camera coordinate system. This transformation consists of a rotation matrix $\mathbf{R} \in SO(3)$ and a translation vector $\mathbf{t} \in \mathbb{R}^3$. 

We are given an object $\mathbf{O}$ with a 3D Scan or CAD model mesh, denoted $\mathcal{M}$, consisting of $N$ 3D vertices $\mathbf{v}_i \in \mathbb{R}^3$, with $i$ being the vertex index. 
A binary encoding of the $N$ vertices, is a binary code $\mathbf{c}_{i}$ of $d$ bits that uniquely corresponds to a vertex $\mathbf{v}_i$. We preprocess the object mesh by upsampling so that $N = 2^d$.

ZebraPose~\cite{su2022zebrapose} constructs this binary encoding iteratively, by splitting the mesh into parts of equal amount of vertices at each step, and assigning a bit to each group. In the iteration $it, it \in \{0, 1, …, d-1\}$ of the surface partition, we have $2^{it}$ separate sub-surfaces. Assuming a surface contains $L$ vertices, balanced k-means is used for the partitioning, resulting in two sub-surfaces containing $\lfloor L/2 \rfloor$ and $L-\lfloor L/2 \rfloor$ vertices respectively. 

This procedure creates a hierarchical encoding, meaning that all vertices whose first $x$ bits are equal, belong to the same sub-surfaces of the object mesh surface until the $x$-th partition. Expressed differently, a binary code $\mathbf{c}$ describes a manifold of coarse-to-fine object surfaces $\mathcal{S}_k, k=0, \cdots, d$ where $\mathcal{S}_{0}$ is the full object mesh and $\mathcal{S}_{d}$ is a single vertex of the mesh.

\subsection{Point-to-Surface Correspondences}{\label{sec:correspond}}
Prior work~\cite{su2022zebrapose} used a binary encoding of surfaces as described in Section~\ref{sec:notation} for pose estimation from RGB images. This was done by training a neural network to estimate a binary code of $d$ bits for every pixel $\mathbf{p}_{u,v}$ within a detected object bounding box. This can then be used to establish 2D-3D correspondences between pixels and encoded vertices $\mathbf{v}_{i}$ of the object model. These correspondences are presented to a Perspective-n-Point (PnP) solver (e.g. RANSAC+EPnP~\cite{lepetit2009ep}) to estimate the object pose. Results show that this encoded surface representation is well suited for neural network training~\cite{su2022zebrapose} since it allows to progressively learn finer details.

However, such a method: (1) Is not designed to make use of depth map information, apart from the final pose refinement stage, (2) does not take explicit advantage of the hierarchical nature of the encoded surface prediction, and (3) does not exploit the inherent confidence in predicted surface codes - instead, the continuous binary code estimates in the range $[0,1]$ are quantized to discrete bit values which discards all confidence information.

In contrast, our method-HiPose is designed to take a single RGB-D image as input and extract features from both modalities to predict 3D-3D correspondences. Our network receives as input a cropped Region of Interest (RoI) from a detected object, both, from the RGB image and the depth map as seen in Figure~\ref{fig:overview}. The inputs are processed by two branches of a bidirectional fusion network as in FFB6D~\cite{he2021ffb6d}. Details on the network architecture and implementation are provided in Section~\ref{subsection:implementation_details}. The pixels of the depth image are converted to a point cloud. For each 3D point $\mathbf{P}$, our network is trained to predict a binary code $\hat{\mathbf{c}}$. This code represents a 3D-3D correspondence between the point cloud and the object model. A solution for the pose can be obtained by passing these correspondences to the Kabsch pose estimation algorithm~\cite{umeyama1991least}.

This approach is in line with similar correspondence methods. However, the hierarchical nature of the encoding presents an opportunity for coarse-to-fine processing that has not been exploited yet. This is where our hierarchical point-to-surface approach is applied. As discussed in Section~\ref{sec:notation}, a binary encoding represents a manifold of object surfaces depending on how many bits of the encoding are considered. Instead of directly processing the full encoding of $d$ bits, which includes high uncertainty for the last bits, we propose to split the binary encoding into two groups: The first $m$ and the last $n$ bits ($d=m+n$). Unlike in ~\cite{su2022zebrapose}, we utilize the ${(m+1)}_{th}$ until the $d_{th}$ bits in an iterative manner, as explained in the following.

The first $m$ bits of the encoding yield a 3D point-to-surface correspondence from a point in the point cloud to a surface segment $\mathcal{S}_{m}$ on the object model. From the surface $\mathcal{S}_{m}$, a centroid point can be computed, and used as a 3D point on the model to create a 3D-3D correspondence. A coarse pose estimate $\hat{\mathbf{R}}_0,\hat{\mathbf{t}}_{0} $ can then be obtained from these correspondences. Crucially, the coarse pose estimate is used for outlier pruning as described in Section~\ref{subsection:pruning}. We iteratively repeat this process for bits $m+1$ until $m+n$, in a coarse-to-fine manner, with the surfaces of interest $\mathcal{S}_{m+1}$ until $\mathcal{S}_{m+n}$ decreasing in size at each iteration. Simultaneously, the pose estimates at each iteration have gradually higher accuracy and can be used for finer outlier removal.     

\subsection{Hierarchical Binary Code Decoding}
\label{subsection:solver}
\begin{figure}[htbp]
\centering
\includegraphics[width=0.45\textwidth]{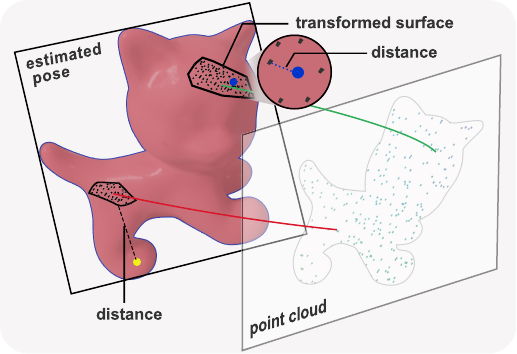}
\caption{\textbf{Correspondence Pruning.} The green line demonstrates an example where the point cloud lies in the transformed surface under the estimated pose. The distance between the point in the point cloud and the transformed surface is represented by the blue dashed line. The red line demonstrates another case where the transformed surface is far away from the point in the point cloud (yellow point). Consequently, the correspondence depicted by the red line will be removed in the next iteration.}
\label{fig:outlier_rejection}
\end{figure}

Existing methods such as ZebraPose~\cite{su2022zebrapose}, use estimated binary codes in a straightforward manner. The continuous estimated code from the neural network is transformed into a binary code by quantizing the values from the range $[0,1]$ to bit values. Then, the estimated binarized code $\hat{\mathbf{c}}$ has direct correspondence to a vertex code $\mathbf{c}_{i}$. However, this method discards valuable confidence information that is inherent in the predicted code and makes the process highly reliant on the performance of the RANSAC-PnP solver.

Denoting the direct (non-binarized) prediction output vector as $\overline{\hat{\mathbf{c}}}$ and its quantized version as ${\hat{\mathbf{c}}}$, we propose to compute a bit correctness probability/confidence vector $\mathbf{p}_{c} \in \mathbb{R}^{d\times1}_{[0,1]}$ as
\begin{equation}
\mathbf{p}_{c} = \mathbf{1}_{d\times1} - |\overline{\hat{\mathbf{c}}} - \hat{\mathbf{c}}|. 
\end{equation} 

HiPose introduces a method to leverage this probability information, enabling superior results to be achieved with several rendering-free iterations, thereby improving algorithmic efficiency. Our binary code decoding consists of initial surface selection and sub-surface partitioning.

\textbf{Initial surface selection.} The blue line in Figure~\ref{fig:overview} indicates the process of initial surface selection. As the number of iterations increases, each surface is further divided and the corresponding bits become more difficult to learn for the network. Therefore, it is essential to select an appropriate starting point $\mathcal{S}_{m}$ from the surface manifold where the pose estimation iterations should begin, or, equivalently select the value $m$, where the binary encoding will be split. 
The initial surface selection for every correspondence is based on the bit correctness probability vector $\mathbf{p}_{c}$.
The last bit $j$, for which $\mathbf{p}_{c}^{j}$ is higher than the probability threshold, is called the trust bit. We set $m_{default}$ as the minimum value of $m$ which limits the maximum initial size of the sub-surfaces and begin our pose estimation iterations at $m = max(j, m_{default})$.
There are two advantages of setting $m_{default}$: Ensuring a certain degree of accuracy in the initial pose estimation and reducing computational complexity.

\textbf{Sub-surface Partitioning.} As shown in Figure~\ref{fig:overview}, the red lines indicate the process of sub-surface partitioning. As the surfaces reduce in size between iterations, correspondences become more challenging to learn and therefore less reliable. Consequently, during each iteration, prior to performing pose estimation on a subdivided surface, we employ our hierarchical correspondence pruning process, which efficiently eliminates outliers within a few iterations, as described in detail in the next Section~\ref{subsection:pruning}.
We perform $d - m$ iterations of hierarchical correspondence pruning and a sub-surface $\mathcal{S}_m$ encoded by $m$ bit can only be partitioned $d - m$ times. If the trust bit $j$ of a prediction is larger than $m_{default}$, this sub-surface $\mathcal{S}_j$ will retain its size in the former $j - m_{default}$ iterations of correspondence pruning. In this way, a prediction with a higher trust bit prioritizes matching a finer correspondence, leading to a more reliable and precise pose in each iteration. For simplicity, we assume that the trust bit $j$ equals $m_{default}$ in Section~\ref{subsection:pruning}, thereby not skipping the sub-surface partitioning.

\subsection{Hierarchical Correspondence Pruning}\label{subsection:pruning}

Through the point-to-surface matching process, starting from bit $m$ of the estimated code, we can compute the centroid $\mathbf{g}_{m}$ of the corresponding surface $\mathcal{S}_{m}$ and all corresponding vertices $\mathbf{v}_i$. The 3D coordinate of $\mathbf{g}_{m}$ is the average of the $M$ vertex coordinates, $\mathbf{g}_m = \frac{1}{M}\sum_{i=1}^{M}\mathbf{v}_i$. 

The sub-surface partitioning and pose estimation is repeated $n$ times (from bit $m + 1$ to bit $d$). In the $it_{\text{th}}$ iteration, $it \in [0,1,...,n-1]$, we compute the centroid $\mathbf{g}_{it}$ of the corresponding sub-surface for every masked point $\mathbf{P}$ in our point cloud and estimate a pose $[\hat{\mathbf{R}}_{it}|\hat{\mathbf{t}}_{it}]$ in this step through a Kabsch solver. With this estimated pose, we select inliers using the distance calculated below.

In the $(it+1)_{\text{th}}$ iteration, we calculate the distance for every correspondence between point $\mathbf{P}$ and transformed sub-surface $\mathcal{S}_{m+it}'$ under pose $[\hat{\mathbf{R}}_{it}|\hat{\mathbf{t}}_{it}]$, which will be used as the threshold to distinguish inliers and outliers. Figure~\ref{fig:outlier_rejection} is a visual representation of this process.
The minimum distance between a point $\mathbf{P}$  in the point cloud and transformed surface $\mathcal{S}_{m+it}'$ is defined as:
\begin{equation}
\label{eq:l}
l = \min\limits_{\mathbf{v}_i} ||\hat{\mathbf{R}}_{it} \mathbf{v}_{i} + \hat{\mathbf{t}}_{it}-\mathbf{P}||
\end{equation}

The correspondences are distinguished as inliers and outliers based on the median of the distance $l$. The different options for distinguishing inliers and outliers are compared in an ablation study, Section~\ref{sec:exp}. Formally, in the general case the pose at iteration $it$ is solved by Kabsch $\mathcal{K}$ with an input set of point-to-surface centroid 3D-3D correspondences as:
\begin{equation}
    \label{eq:posekabsch}
    [\hat{\mathbf{R}}_{it}|\hat{\mathbf{t}}_{it}] = \mathcal{K}(\{\mathbf{P}_{k}, \mathbf{g}_{m+it}^{k}\}, {\mathbf{P}_{k}\in \text{inliers}_{it}}
\end{equation}

After completing the $n$ iterations of surface partitioning, all point-to-surface correspondences have converged to point-to-point correspondences. Finally, we perform one round of the Kabsch algorithm with all the point-to-point correspondences that were never recognized as outliers in the hierarchical correspondence pruning process to generate the final estimated pose, 
$[\hat{\mathbf{R}}_{n}|\hat{\mathbf{t}}_{n}]$.

%% file: Sections/4_exp.tex
\section{Experiments}
\label{sec:exp}

In this section, we start with the implementation, datasets and evaluation metrics. Next, we show ablation studies of our method with different 3D-3D correspondence solvers. Finally, we present the experimental results of our method and compare to recent pose estimation methods from the literature and from the BOP Challenge.

\subsection{Implementation Details.}
\label{subsection:implementation_details}
Our approach can be easily integrated into a variety of existing RGB-D networks. This paper utilizes the full flow bidirectional fuse network from FFB6D~\cite{he2021ffb6d} as a baseline and applies some modifications to create the HiPose network.

Similarly to FFB6D, our HiPose network has two branches to deal with images and point clouds, respectively. We feed the HiPose network with 1) a zoomed-in Region of Interest (RoI) image, and 2) a point cloud generated by uniform sampling of a fixed number of pixels from the RoI depth map.

Modifications are also made to the output layers. We replace three output heads with a single head containing the visibility mask and a binary encoding of length $d$ for every randomly selected point. We use L1 loss for both the visible mask and binary encoding. Our training loss is defined as
\begin{equation}
Loss = L_{mask} + \alpha L_{code}   
\end{equation}
where $\alpha$ is a weight factor between the mask and the binary encoding losses, set to $\alpha = 3$ throughout all experiments. Note that for the binary encoding prediction, we only calculate a loss for the points within the predicted visible mask.

For the network backbone, we use the recent ConvNext~\cite{Liu2022ACF} architecture which builds up on ResNet~\cite{He2015DeepRL}. ConvNext shows on-par performance to Vision Transformer~\cite{Dosovitskiy2020AnII} while retaining the efficiency and simplicity of ResNet. 

We train our network for $380K$ iterations with a batch size of $32$. We use the Adam~\cite{Kingma2014AdamAM} optimizer with a fixed learning rate of $1e-4$. During training, we employ RGB image augmentations used in ZebraPose~\cite{su2022zebrapose} and depth augmentations from MegaPose~\cite{Labbe2022MegaPose6P} for the depth maps.

\subsection{Datasets}
We conduct our experiments on the LM-O~\cite{Brachmann2016UncertaintyDriven6P}, YCB-V~\cite{xiang2018posecnn} and T-LESS~\cite{hodan2017t} datasets. These datasets collectively encompass a wide range of scenarios, including instances of heavy occlusion, texture deficiency and symmetrical objects. 
Since annotating real data for object pose can be very time-consuming, we utilize publicly available synthetic physically-based rendered (PBR) images provided by the BOP challenge~\cite{Sundermeyer2023BOPC2} to demonstrate that our network can be effectively trained using only synthetic data.

\subsection{Metrics}

For the LM-O dataset, we report the ADD(-S) metric which is the most common metric for 6DoF pose among contemporary works. ADD calculates the percentage of object vertices that fall under a distance threshold (object size dependent) when projected to the camera coordinates using the estimated pose vs. using the ground truth pose.
In the case of a symmetric object, ADD(-S) differs in that it matches the closest model points (taking symmetry into account) instead of the exact same model points.
For the YCB-V dataset, we report the area under curve (AUC) of the ADD(-S) with a maximum threshold of 10cm as described in~\cite{xiang2018posecnn}. Additionally, for both datasets, we report the BOP score metric defined by the BOP Challenge~\cite{Sundermeyer2023BOPC2}.

\subsection{Ablation Studies}
\input{Tables/Tab_ablation}
\input{Tables/Tab_outlier_removal_metrics}
\input{Sections/line_chart}

\input{Tables/Tab_LMO_ADD}

\input{Tables/Tab_YCBV_AUC}
\input{Tables/Tab_BOP_score}

In the following, we perform several ablation studies with the LM-O dataset. We summarize the results in Table~\ref{tab:ablation_study}, Table~\ref{tab:ourlier_removal_metrics} and  Figure~\ref{fig:ablation_initial_bit}.

\textbf{Effectiveness of Correspondence Pruning.}
We first focus on the effectiveness of our proposed hierarchical correspondence pruning.  
In the experiment (A0) in Table~\ref{tab:ablation_study}, we directly solve the object pose with the Kabsch Algorithm. The promising results highlight the effectiveness of the binary encoding. However, compared to our other experiments reveals that the predicted correspondences from the network are still noisy and contain outliers. 

Comparing A1 with A0, the most common method for identifying outliers using RANSAC framework, we observed a $2.47\%$ recall improvement. The results of A1 heavily depend on the choice of hyper-parameters, including the number of correspondences used in each iteration, the number of RANSAC iterations, and the inlier threshold in each iteration. Additionally, random seed variations can also impact the results. In this experiment, we utilized the public RANSAC and Kabsch algorithm from Open3D~\cite{zhou2018open3d}. Note that we explored multiple parameter combinations and reported the best results among them.

In contrast to the RANSAC scheme, our hierarchical correspondence pruning provides stable results analytically, irrespective of the random seed. 
In experiment A2, we chose the $10_\text{th}$ bit as our initial bit and defined the confidence bit based on predicted logits higher than $0.52$ or lower than $0.48$. As shown in Table~\ref{tab:ablation_study}, compared to not using any outlier strategy (A0), our approach (A2)  improves recall by about $3\%$ while outperforming the best results achievable by RANSAC. 

We also estimate the precision metrics of outlier removal at each iteration in Table~\ref{tab:ourlier_removal_metrics}. A true sample is defined when the distance between estimated coordinates and ground truth fall under a 10mm threshold. The increase in precision confirms the gradual removal of low-quality correspondences. In the following ablation studies, we demonstrate that our approach is also robust to the choice of hyper-parameters.

\textbf{Influence of Default Initial Bit.} Using smaller initial bits implies matching the point to a relatively coarse surface correspondence. However, our trust-bit-based initial surface selection ensures that each vertex is considered separately and corresponds to its most reliable initial bit.

As demonstrated in Figure~\ref{fig:ablation_initial_bit}, our proposed design is robust to the choice of initial bit from $5_\text{th}$ bit to $11_\text{th}$ bit. We observe some performance drops when we start from the $12_\text{th}$ bit, and a significant drop when directly using $16_\text{th}$ bit, underscoring the importance of our hierarchical correspondence pruning. By calculating the mean recall across all 8 objects, we noticed that using the $9_\text{th}$ to $11_\text{th}$ bits as the initial bit provides slightly higher results. Considering both accuracy and computational efficiency, we consistently use the $10_\text{th}$ bit as our default initial bit.

\textbf{Threshold for the Trust Bit.}
The initial bit for each point and our inlier identification strategy is closely tied to the choice of the trust bit. By varying the threshold $0.02$ used in experiment A2 (predicted logits greater than 0.52 or smaller than 0.48), we alter the trust bit threshold in experiments B0, B1, and B2. As indicated by the experimental results, when the threshold is greater than $0.08$, we start to observe a small performance drop. Overall, the results remain quite stable within the threshold range of $0.02$ to $0.08$. This demonstrates the relative robustness of our approach to the choice of the trust bit.

\textbf{Criteria used in Correspondence Pruning.}
The default criterion for distinguishing inliers and outliers in correspondence pruning is based on the median of distance $l$ defined in Equation~\ref{eq:l}. We replaced the median criterion for the inlier threshold used in A2 with a mean criterion in C0, resulting in a decrease in average recall. It is comprehensible that utilizing indicators associated with the median produces superior outcomes, given the median's ability to disregard the impact of extreme values.

\textbf{Effectiveness of CNN Backbone.}
To ensure comparability with recent research employing the transformer architecture and ConvNext~\cite{Liu2022ACF} feature backbone, we default to using ConvNext as our image feature extraction network. Additionally, we provide results of experiment (D0) in Table~\ref{tab:ablation_study} using ResNet as the feature backbone to offer further insights for comparison with earlier approaches. Results show that the choice of feature backbone only has a marginal effect. 

\textbf{Naive RGB-D baselines.}
Using the networks provided by ZebraPose, we back-project 2D pixels into 3D points using the depth map, followed by pose estimation using RANSAC + Kabsch for 3D-3D correspondence(E0). We also implement "Concat"(E1), a naive baseline model that learns 2D-3D correspondence, yet solves the pose with 3D-3D correspondences. In the "Concat" baseline, we concatenate the RGB and depth channels as input for the CNN. However, the absence of a pretrained CNN model appears to make the results worse. Nonetheless, none of the baselines surpass HiPose, suggesting that direct learning of 3D-3D correspondences is more effective.

\subsection{Comparison to State of the Art}
In the following, we compare HiPose to the state of the art using various metrics on multiple datasets.

\textbf{Results on LM-O.} In Table~\ref{tab:lmo_results_table_}, we compare our HiPose with the state-of-the-art methods on the LM-O dataset w.r.t. the ADD(-S) score. We used the 2D detection provided by CDPN~\cite{li2019cdpn}, which is based on the FCOS~\cite{Tian2019FCOSFC} detector and trained only with PBR images provided by the BOP challenge. According to the results, HiPose outperforms all other methods by a large margin of $11.9\%$ compared to the best RGB-D method DFTr and $12.7\%$ compared to the best RGB-only method ZebraPose.

\textbf{Results on YCB-V.} In Table~\ref{tab:ycbv_results_table}, we compare HiPose with the state-of-the-art methods on the YCB-V dataset w.r.t. the AUC of ADD-S and ADD(-S) score. All other methods used real and synthetic images in the training. To ensure a fair comparison, the 2D FCOS detections employed here are trained with both real and synthetic images. According to the results, HiPose again excels beyond all other methods with a significant margin (around $1\%$) when taking into account that scores on YCB-V are already close to saturation.

\textbf{Results on the BOP Benchmark.} The BOP benchmark provides a fairer ground for comparisons, offering uniform training datasets and 2D detections for all participating methods and more informative evaluation metrics~\cite{hodan2018bop}. We used the default detections provided for the BOP Challenge 2023.

Most methods in Table~\ref{tab:bop_score_table} rely on a time-consuming pose refinement step, while HiPose estimates accurate object pose directly without any pose refinement. HiPose surpasses the state-of-the-art on LM-O, YCB-V datasets and achieves a very comparable result on the T-LESS dataset. When considering the average recall across the three datasets, HiPose exhibits higher recall compared to all other methods.

GDRNPP~\cite{liu2022gdrnpp_bop} has the most closely aligned results with HiPose, yet HiPose is approximately $\mathbf{40}$ times faster than GDRNPP with refinement. This demonstrates that HiPose is both accurate and computationally efficient.

\subsection{Runtime Analysis}
The inference time mainly comprises two components: 1) object detection and 2) object pose estimation. For a fair comparison, we use the same method to calculate inference speed as GDRNPP on an RTX3090 GPU. Our average object pose estimation time across the LM-O, YCB-V, and T-LESS datasets is $0.075$ seconds per image. The average 2D detection time with YOLOX~\cite{Ge2021YOLOXEY} on those three datasets is $0.081$ seconds. The fast object pose estimation time ensures the real-time applicability of our approach, especially since the costly refinement methods are not necessary.

%% file: Tables/Tab_ablation.tex
\begin{table}
  \centering
  \begin{tabular}{@{}c|c|c@{}}
    \toprule
     A0 &Kabsch& 86.65 \\
     A1 &RANSAC+ Kabsch& 89.12 \\
     A2 &Our Hierarchical Pruning &\textbf{89.62}\\
     \midrule
     B0 &Trust Bit Threshold 0.08& 89.55 \\
     B1 &Trust Bit Threshold 0.06& 89.61 \\
     B2 &Trust Bit Threshold 0.04& 89.59 \\
     \midrule
     C0 &Inlier Threshold: median $\rightarrow$ mean & 89.47 \\
     \midrule
     D0 & ConvNext-B~\cite{Liu2022ACF} $\rightarrow$ ResNet34~\cite{He2015DeepRL} & 88.64 \\
     \midrule
     E0 & ZebraPose(pbr)+RANSAC Kabsch&87.0\\
     E1 & ConcatRGBD+RANSAC Kabsch&84.5\\
    \bottomrule
  \end{tabular}
  \caption{\textbf{Ablation study on LM-O~\cite{Brachmann2016UncertaintyDriven6P}.} We conduct several ablation studies, comparing our proposed method to a RANSAC-based approach and exploring the impact of hyperparameters on the results. The results are presented in terms of average recall of ADD(-S) in \%.}
  \label{tab:ablation_study}
\end{table}

%% file: Tables/Tab_outlier_removal_metrics.tex
\begin{table}[t]
  \centering
  \scalebox{0.78}{
  \begin{tabular}{c|c|c|c|c|c|c|c|c}
    \toprule
    Iteration Step & 1 & 2 & 3 & 4 & 5 & 6 & 7 & final \\
    \midrule
    precision(\%) & 66.1 &67.4 &70.3 &73.3 &73.8 &75.9 &75.9 &76.1 \\
    \bottomrule
  \end{tabular}
  }
  \caption{\textbf{Outlier removal metrics.} We evaluate precision at each iteration to validate the outlier removal process.}
  \label{tab:ourlier_removal_metrics}
\end{table}

%% file: Sections/line_chart.tex
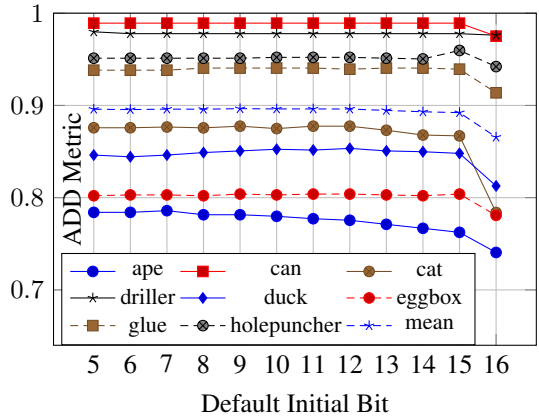
\begin{figure}[t]
\begin{tikzpicture}
\begin{axis}[
scaled y ticks=real:100,
ytick scale label code/.code={},
ymin = 64,
ymax = 100,
symbolic x coords={5, 6, 7, 8, 9, 10, 11, 12, 13, 14, 15, 16},
xtick=data,
height=6cm,
width=8cm,
grid=major,
xlabel={Default Initial Bit},
ylabel={ADD Metric},
y label style={at={(0.23,0.5)}},
legend style={
        at={(0.445,0.0)},
        anchor=south,
        legend columns=3,
        nodes={scale=0.85, transform shape}
    },
]

\addplot coordinates {
(5,78.41) 
(6, 78.41) 
(7, 78.59) 
(8, 78.15) 
(9, 78.15) 
(10, 77.98)
(11, 77.72)
(12, 77.55)
(13, 77.11)
(14, 76.68)
(15, 76.24)
(16, 74.06)           
};

\addplot coordinates {
(5,98.92) 
(6, 98.92) 
(7, 98.92) 
(8, 98.92) 
(9, 98.92)
(10, 98.92)
(11, 98.92)
(12, 98.92)
(13, 98.92)
(14, 98.92)
(15, 98.92)
(16, 97.51)
};

\addplot coordinates {
(5,87.58) 
(6, 87.58) 
(7, 87.66) 
(8, 87.57) 
(9, 87.75)
(10, 87.49)
(11, 87.75)
(12, 87.75)
(13, 87.31)
(14, 86.79)
(15, 86.7)
(16, 78.39)
};

\addplot coordinates {
(5,97.98) 
(6, 97.78) 
(7, 97.78) 
(8, 97.78) 
(9, 97.78)
(10, 97.78)
(11, 97.78)
(12, 97.78)
(13, 97.78)
(14, 97.78)
(15, 97.78)
(16, 97.61)
};

\addplot coordinates {
(5,84.62) 
(6, 84.43) 
(7, 84.62) 
(8, 84.89) 
(9, 85.07)
(10, 85.25)
(11, 85.16)
(12, 85.34)
(13, 85.07)
(14, 84.98)
(15, 84.8)
(16, 81.27)
};

\addplot coordinates {
(5,80.21) 
(6, 80.3) 
(7, 80.3) 
(8, 80.21) 
(9, 80.39)
(10, 80.3)
(11, 80.39)
(12, 80.39)
(13, 80.3)
(14, 80.21)
(15, 80.39)
(16, 78.08)
};

\addplot coordinates {
(5,93.82) 
(6, 93.82) 
(7, 93.82) 
(8, 94.05) 
(9, 94.05)
(10, 94.05)
(11, 94.05)
(12, 93.93)
(13, 94.05)
(14, 94.05)
(15, 93.93)
(16, 91.37)
};

\addplot coordinates {
(5,95.12) 
(6, 95.12) 
(7, 95.12) 
(8, 95.12) 
(9, 95.12)
(10, 95.21)
(11, 95.21)
(12, 95.21)
(13, 95.12)
(14, 95.04)
(15, 95.96)
(16, 94.21)
};

\addplot coordinates {
(5,89.59) 
(6, 89.55) 
(7, 89.60) 
(8, 89.59) 
(9, 89.65)
(10, 89.62)
(11, 89.62)
(12, 89.61)
(13, 89.46)
(14, 89.31)
(15, 89.22)
(16, 86.56)
};

\legend{ape, can, cat, driller, duck, eggbox, glue, holepuncher, mean}
\end{axis}
\end{tikzpicture}
\vspace{-0.3cm}
     \caption{We conduct an ablation study on selecting the default initial bit $m_{default}$ using $8$ objects from the LM-O dataset.
     The flat curve illustrates that our proposed design is robust and has a clear advantage to the non-hierarchical variant ($16$ as initial bit). }
     \label{fig:ablation_initial_bit}
\end{figure}

%% file: Tables/Tab_LMO_ADD.tex
\begin{table*}
  \centering
  \begin{tabular}{c|c|c|c|c|c|c|c}
    \toprule 
    \multirow{2}{*}{Method} & \multicolumn{2}{c|}{RGB Input} & \multicolumn{5}{c}{RGB-D Input} \\ 
         \cline{2-8}
      & GDR-Net~\cite{wang2021gdr} 
& ZebraPose~\cite{su2022zebrapose}&  PR-GCN~\cite{Zhou2021PRGCNAD} 
&  FFB6D~\cite{he2021ffb6d} 
& RCVPose~\cite{wu2022vote} 
& DFTr~\cite{zhou2023deep} &\textbf{Ours}\\
    \midrule
     ape & 46.8 
& 57.9& 40.2 
& 47.2 
&  60.3 
& 64.1 &  \textbf{78.0}\\
     can & 90.8 
& 95.0& 76.2 
& 85.2 
&  92.5 
& 96.1 &  \textbf{98.9}  \\
     cat & 40.5 
& 60.6& 57.0 
& 45.7 
&  50.2 
& 52.2 &  \textbf{87.5}\\
     driller & 82.6 
& 94.8& 83.2 
& 81.4 
&  78.2 
& 95.8 &  \textbf{97.8}\\
     duck & 46.9 
& 64.5& 30.0 
& 53.9 
&  52.1 
& 72.3 &  \textbf{85.3}\\
     eggbox* & 54.2 
& 70.9& 68.2 
& 70.2 
&  \textbf{81.2} 
& 75.3 &  80.3\\
     glue* & 75.8 
& 88.7& 67.0 
& 60.1 
&  72.1 
& 79.3 &  \textbf{94.1}\\
     holepuncher & 60.1 
& 83.0& \textbf{97.2} 
& 85.9 
& 75.2 
& 86.8 &  95.2  \\
     \hline
     mean & 62.2 & 76.9&  65.0 &  66.2 &  70.2 & 77.7 & \textbf{89.6}\\
    \bottomrule
  \end{tabular}
  \caption{\textbf{Comparison with State of the Art on LM-O~\cite{Brachmann2016UncertaintyDriven6P}}. We report the Recall of ADD(-S) in \% and compare with state of the art. (*) denotes symmetric objects.}
  \label{tab:lmo_results_table_}
\end{table*}

%% file: Tables/Tab_YCBV_AUC.tex
\begin{table}
  \centering
  \begin{tabular}{@{}l|c|c|c@{}}
    \toprule
     \multicolumn{2}{c|}{Method} & \makecell{AUC of\\ADD-S} & \makecell{AUC of\\ADD(-S)} \\
    \midrule
    \multirow{3}{*}{\rotatebox{90}{RGB}} & SO-Pose~\cite{di2021so} &  90.9 &  83.9  \\
     & GDR-Net~\cite{wang2021gdr} &  91.6 &  84.4 \\
     & ZebraPose~\cite{su2022zebrapose} & 90.1 & 85.3\\
     \midrule
    \multirow{5}{*}{\rotatebox{90}{RGB-D}} & PVN3D~\cite{he2020pvn3d} &  95.5 &  91.8 \\
     & RCVPose~\cite{wu2022vote} &  96.6 &  95.2 \\
     & FFB6D~\cite{he2021ffb6d} &  96.6 &  92.7 \\
     & DFTr~\cite{zhou2023deep} & 96.7 & 94.4\\
     & Ours &  \textbf{97.6} &  \textbf{95.5}  \\
    \bottomrule
  \end{tabular}
  \caption{\textbf{Comparison with State of the Art on YCB-V\cite{xiang2018posecnn}}. We compare our HiPose with state of the art w.r.t AUC of ADD(-S) and AUC of ADD-S in \%.}
  \label{tab:ycbv_results_table}
\end{table}


%% file: Tables/Tab_BOP_score.tex
\begin{table*}
  \centering
  \begin{tabular}{l|c|c|c|c|c|c}
    \toprule
     \textbf{Method} & Refinement & LM-O & YCB-V & T-LESS & mean & time(sec) \\
    \midrule
    FFB6D-CVPR21-PBR-NoRefinement~\cite{he2021ffb6d} &  - & 68.7 &  75.8  & - & 72.3$^*$ & 0.19$^*$\\
    RCVPose 3D\_SingleModel\_VIVO\_PBR~\cite{wu2022vote} &  ICP~\cite{Rusinkiewicz2001EfficientVO} & 72.9 &  84.3  & 70.8 & 76.0 & 1.33\\
    SurfEmb-PBR-RGBD~\cite{Haugaard2021SurfEmbDA} &  custom~\cite{Haugaard2021SurfEmbDA} & 76.0 &  79.9  & 82.8 & 79.6 & 9.04\\
    RADet+PFA-PBR-RGBD~\cite{Hai2023RigidityAwareDF} &  PFA~\cite{hu2022perspective} & 79.7 &  82.6  & 80.2 & 80.8 & 2.63\\
    GDRNPP-PBR-RGBD-MModel~\cite{liu2022gdrnpp_bop} &  $\sim$CIR~\cite{lipson2022coupled} & 77.5 &  90.6  & \textbf{85.2} & 84.4 & 6.37\\
    HiPose (ours) &  - & \textbf{79.9} &  \textbf{90.7}   & 83.3 & \textbf{84.6} & \textbf{0.16}\\
    \bottomrule
  \end{tabular}
  \caption{Comparison to leading methods of the BOP Challenge~\cite{Sundermeyer2023BOPC2} that trained on synthetic PBR data only w.r.t. BOP score.  ($\sim$) denotes similar to CIR\cite{lipson2022coupled}. $^*$ averaged over LM-O and YCB-V only as T-LESS results are not provided for this method.}
  \label{tab:bop_score_table}
\end{table*}


%% file: Sections/5_conclusion.tex
\section{Conclusion}
We introduced HiPose, an RGB-D based object pose estimation method designed to fully exploit hierarchical surface encoding representations and iteratively establish robust correspondences. In contrast to existing methods, we consider the confidence of every bit prediction and gradually remove outliers. Our method is trained exclusively on synthetic images and outperforms the state-of-the-art in object pose estimation accuracy across several datasets. At the same time, HiPose is considerably faster as time-consuming pose refinements become redundant.

%% file: Sections/sup.tex
\begin{figure}[t]
\centering
\includegraphics[width=0.5\textwidth]{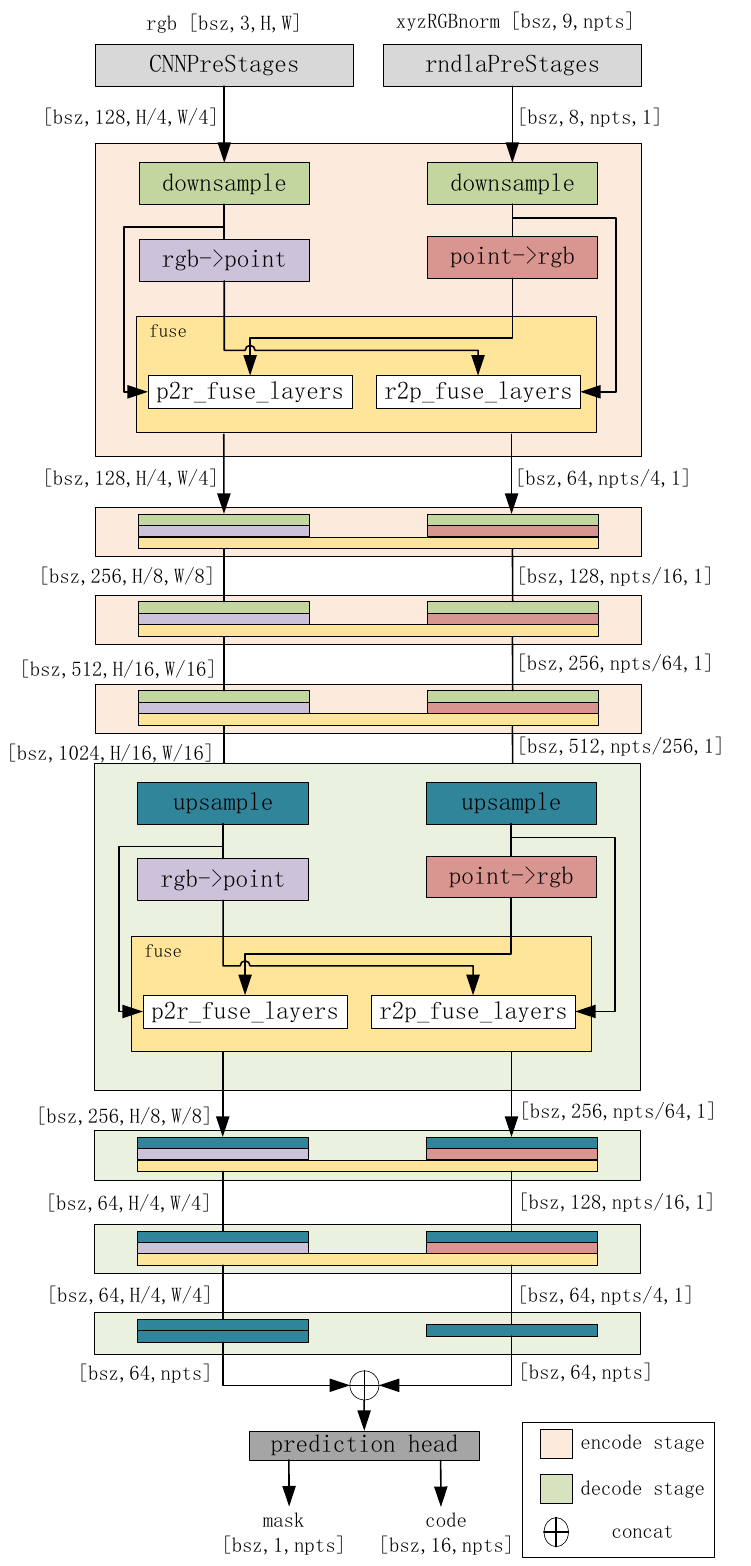}
\caption{\textbf{Network Architecture :} The network comprises four encoder blocks and four decoder blocks. Each block performs upsampling or downsampling of the input, processes the RGB and point features, and subsequently merges them except the last decoder block. In the RGB image branch, we employ ConvNeXt blocks~\cite{Liu2022ACF} as the encoders and PSPNet blocks~\cite{zhao2017pyramid} as the decoders. As for the point cloud branch, we utilize modules derived from Randla~\cite{hu2020randla}. Here, 'bsz' refers to the batch size, 'npts' denotes the number of points, and 'H/W' represents the height and width of the image.}
\label{fig:detail_network}
\end{figure}

\begin{figure}[t]
\centering
\includegraphics[width=0.4\textwidth]{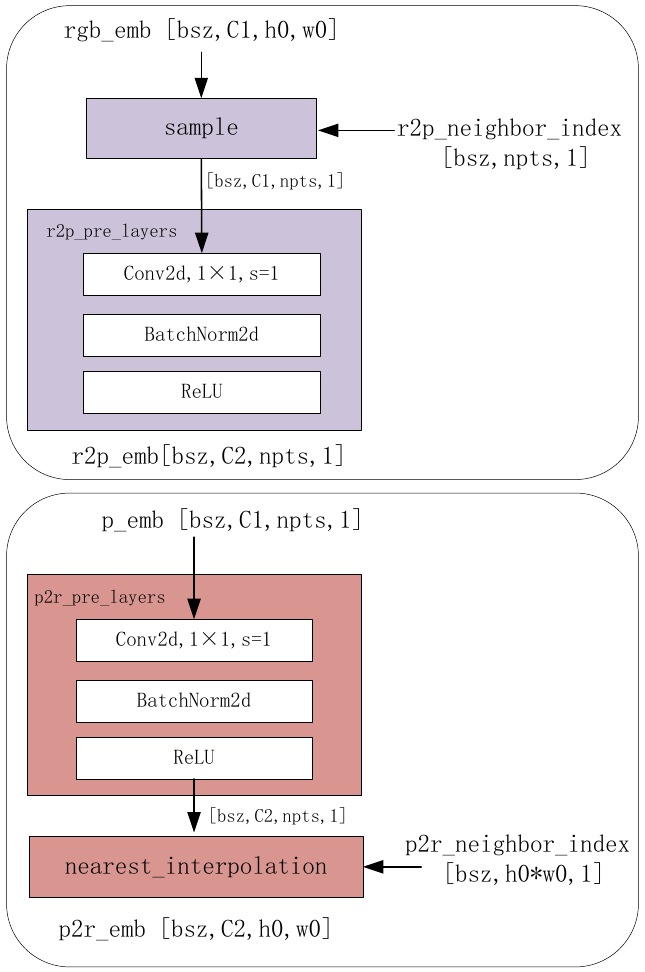}
\caption{The upper block represents the conversion of RGB features to point features, denoted by $r2p\_emb$, while the bottom block illustrates the conversion of point features to RGB features, denoted by $p2r\_emb$. $r2p\_neighbor\_index$  indicates the index of the closest pixel for each point feature, similar for $p2r\_neighbor\_index$.}
\label{fig:p2r_r2p}
\end{figure}

\begin{figure}[t]
\centering
\includegraphics[width=0.4\textwidth]{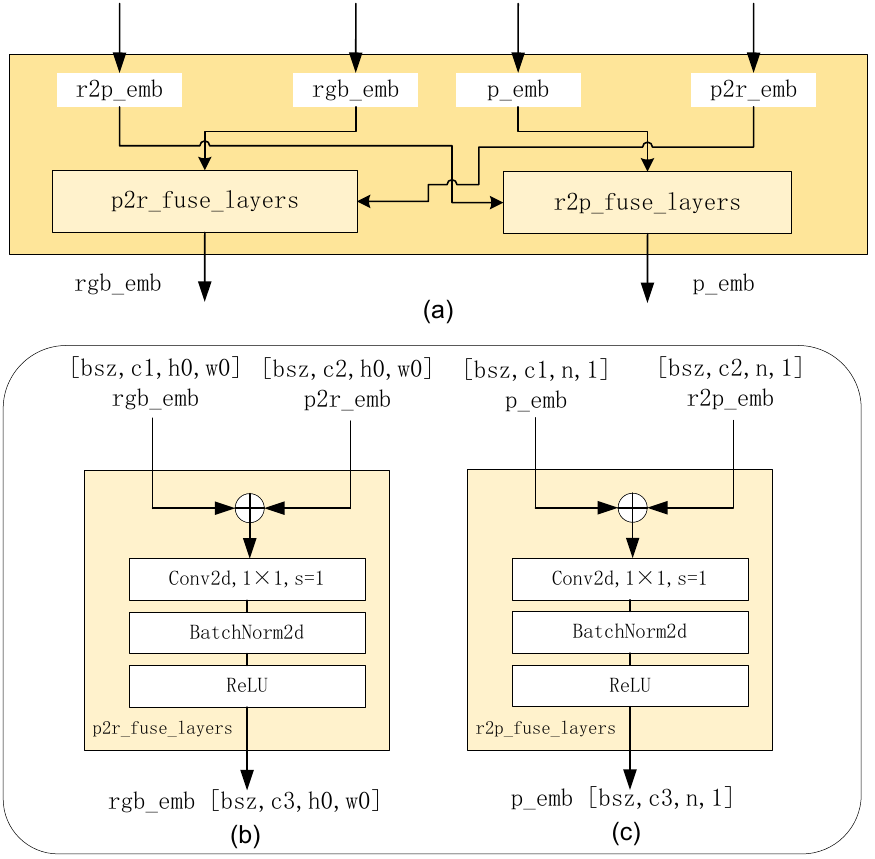}
\caption{(a) The feature flow of the fuse block. (b) This block serves the purpose of fusing RGB feature $rgb\_emb$ and point-to-RGB feature $p2r\_emb$. (c) This block is responsible for fusing point feature $p_emb$ and RGB-to-point feature $r2p\_emb$.}
\label{fig:fuse}
\end{figure}

\subsection{Details of Network}
Figure~\ref{fig:detail_network} illustrates the architecture of HiPose. The network comprises two branches, namely the RGB branch and the point cloud branch. In the pre-stage of the RGB branch, a cropped image with dimensions of $3 \times H \times W$ is transformed into RGB embedding with dimensions of $128 \times H/4 \times W/4$. Here, $H$ and $W$ represent the height and width of the input cropped RGB image, respectively, and by default, both are set to 256. On the other hand, the point cloud branch maps an input with 9 channels, consisting of point coordinates, color, and normal information, to a feature. The parameter $npts$ is set to 2730, which denotes the number of randomly sampled valid input point clouds.

Each of these branches is constructed with interconnected encoders and decoders, serving the fundamental purpose of feature extraction, feature transformation, and feature fusion.

The process of feature extraction aims to extract high-level features and adjust the channel dimension. In accordance with FFB6D~\cite{he2021ffb6d}, RandLA-Net~\cite{hu2020randla} is employed to handle point cloud features. Furthermore, pre-trained ConvNeXt-B~\cite{Liu2022ACF} and PSPNet~\cite{zhao2017pyramid} models are incorporated into the encoder and decoder blocks.

Feature transformation refers to the conversion between the features of the RGB branch and the point cloud branch, facilitated through coordinate correspondence. Specifically, as demonstrated in Figure~\ref{fig:p2r_r2p}, the point cloud branch feature can be generated by aggregating features from the nearest features in the RGB branch. Likewise, the RGB branch feature can be generated by interpolating the feature from the point cloud branch. This enables bidirectional transformation between the features of the RGB branch and the point cloud branch.

The process of feature fusion is executed using a Convolutional Neural Network (CNN). The new RGB feature is generated by concatenating the RGB feature with the transformed RGB feature, and the same procedure is applied to the depth feature. Further details regarding the feature fusion process can be observed in Figure~\ref{fig:fuse}.

Finally, a straightforward convolution-based head is employed to predict the visible mask and code for the selected $npts$ points.

\subsection{Details of Open3D RANSAC+Kabsch}
\begin{table}
  \centering
  \scalebox{0.95}{
  \begin{tabular}{@{}c|c|c|c|c|c|c@{}}
    \toprule
     \# points  & 4 & 6 & 8 & 10 & 20 & 30 \\
     \midrule
     500 iterations & 88.7 & 89.0 & 89.0	& \textbf{89.1} & 89 & 88.7 \\
     1000 iterations & 89.0 & \textbf{89.1} & 88.9 & \textbf{89.1} & \textbf{89.1} & 88.8 \\
     1500 iterations & 89.0 & \textbf{89.1} & \textbf{89.1} & \textbf{89.1} & 88.9 & 88.8 \\
    \bottomrule
  \end{tabular}
  }
  \caption{\textbf{Test RANSAC+Kabsch parameters on LM-O~\cite{Brachmann2016UncertaintyDriven6P}.} We tune the number of correspondences in each RANSAC iteration and the number of RANSAC iterations with a maximum correspondence points-pair distance of 2cm. The results are presented with average recall of ADD(-S) in \%. According to the table, using 10 correspondences in each RANSAC iteration yields the best results. However, the results achievable with RANSAC+Kabsch are inferior to those obtained with our hierarchical approach.}
  \label{tab:ablation_study_RANSAC_parameters}
\end{table}

We use the $registration\_ransac\_based\_on\_correspondence$ function in Open3D\cite{zhou2018open3d} to solve the object pose with the given correspondence. We tuned the number of correspondences in each RANSAC iteration and the number of RANSAC iterations. The results achievable with RANSAC+Kabsch are inferior to those obtained with our hierarchical approach, as showed in Table.~\ref{tab:ablation_study_RANSAC_parameters}.

\subsection{Impact of ICP}
\begin{table}[h]
  \centering
  \scalebox{0.75}{
  \begin{tabular}{@{}l|c@{}}
    \toprule
    Experiment Setup & ADD(-S) in \% \\
    \midrule
    ZebraPose (Trained only with pbr images) & 63.5\\
    ZebraPose (pbr) + ICP & 83.9\\
    ZebraPose (pbr) + RANSAC Kabsch & 87.0\\
    ZebraPose (pbr) + RANSAC Kabsch + ICP & 87.0\\
    \midrule
    \textbf{HiPose} (ours) &  \textbf{89.6}\\
    HiPose + ICP refinement (with ground truth object mask) &  89.3\\
    \bottomrule
  \end{tabular}
  }
  \caption{\textbf{Evaluate the impact of ICP.} We assess the impact of ICP on both HiPose and ZebraPose, both of which are trained solely with pbr images. We observed that once the pose achieves a satisfactory level of accuracy, the incorporation of ICP does not lead to betterment.}
  \label{tab:influence_of_icp}
\end{table}

The Iterative Closest Point algorithm (ICP) is commonly employed as a refinement strategy, leveraging depth information to align the estimated pose. We assess the impact of ICP on both HiPose and ZebraPose, both of which are trained solely with pbr images in Table.~\ref{tab:influence_of_icp}. For HiPose, we provide a ground truth object mask to facilitate the application of ICP. Surprisingly, ICP fails to yield any enhancements and, in fact, worsens the outcome. In the case of ZebraPose, a substantial improvement in the result is observed. Nevertheless, once the pose achieves a satisfactory level of accuracy, such as employing RANSAC Kabsch (recall greater than $87\%$), the incorporation of ICP does not lead to betterment. This circumstance may be attributed to insufficient accuracy in the depth map.

\subsection{Impact of noisy depth}
\begin{table}[h]
  \centering
  \scalebox{0.75}{
      \begin{tabular}{@{}l|c@{}}
        \toprule
        Experiment Setup & ADD(-S) in \% \\
        \midrule
        \textbf{HiPose} (ours) &  \textbf{89.6}\\
        \midrule
        Depth with Zero Mean Gaussian Noise with Sigma 0.01 & 89.0\\
        Random drop $20\%$ points in Depth Map & 89.5 \\
        \bottomrule
      \end{tabular}
  }
  \caption{\textbf{Evaluate the impact of noisy depth.} When introducing noise or randomly omitting data points in the depth map, HiPose still performs admirably under such circumstances.}
  \label{tab:influence_of_noisy_depth}
\end{table}

During training, we augmented the depth maps with Gaussian noise and randomly dropped pixels, to make the network less sensitive to the noise. Coincidentally, the 3 evaluated datasets are captured with different depth sensors, showing that HiPose is robust to different noise levels. We perform additional experiments in Table.~\ref{tab:influence_of_noisy_depth}, showing that HiPose is quite robust to missing measurements in the depth map. However, inaccurate measurements do slightly affect performance.

\subsection{Details of YCB-V results}
We summarized the per-object results on the YCB-V dataset~\cite{xiang2018posecnn} in Table~\ref{tab:ycbv_full_results_AUC}. As presented in the table, we outperform other approaches on most test objects.

\begin{table*}[ht]
  \centering
  \begin{tabular}{@{}l|c|c|c|c|c|c|c|c@{}}
    \toprule 
     Method & \multicolumn{2}{c|}{PVN3D~\cite{he2020pvn3d}} & \multicolumn{2}{c|}{FFB6D~\cite{he2021ffb6d}} &  \multicolumn{2}{c|}{DFTr~\cite{zhou2023deep}} &  \multicolumn{2}{c}{\textbf{Ours}}\\
    \midrule
    Metric & \makecell{AUC of \\ADD-S} & \makecell{AUC of \\ADD(-S)} &\makecell{AUC of \\ADD-S} & \makecell{AUC of \\ADD(-S)} & \makecell{AUC of \\ADD-S} & \makecell{AUC of \\ADD(-S)} & \makecell{AUC of \\ADD-S} & \makecell{AUC of \\ADD(-S)} \\ 
    \midrule
     002\_master\_chef\_can & 96.0&  80.5&  96.3& 80.6& \textbf{97.0}& \textbf{92.3}      & 96.4	&86.2\\
     003\_cracker\_box      & 96.1&  94.8&  96.3& 94.6 &95.9 &93.9         & \textbf{97.7}&	\textbf{96.7}\\
     004\_sugar\_box        & 97.4&  96.3&  97.6& 96.6& 97.1& 95.5       & \textbf{98.2}&	\textbf{97.1}\\
     005\_tomato\_soup\_can & 96.2&  88.5 & 95.6& 89.6& 95.6& 92.6      & \textbf{97.0}	&\textbf{95.1}\\
     006\_mustard\_bottle   & 97.5&  96.2 &  97.8& \textbf{97.0}& 97.6& 96.3           & \textbf{98.4}&	96.9\\
     007\_tuna\_fish\_can   & 96.0&  89.3& 96.8 &88.9 &97.3 &94.5           & \textbf{97.8}&	\textbf{96.2}\\
     008\_pudding\_box      & 97.1 &  95.7& 97.1 &94.6 &97.4 &95.7       & \textbf{98.8}&	\textbf{98.1}\\
     009\_gelatin\_box      & 97.7&  96.1&  98.1 &96.9 &97.6& 96.3           & \textbf{98.9}	&\textbf{97.8}\\
     010\_potted\_meat\_can & 93.3 &  88.6&  94.7& 88.1& \textbf{95.9}& \textbf{92.1}         & 93.5	&83.4\\
     011\_banana            & 96.6 &  93.7&  97.2 &94.9 &97.1 &95.0          & \textbf{98.6}&	\textbf{96.3}\\
     019\_pitcher\_base     & 97.4&  96.5& \textbf{97.6}& \textbf{96.9}& 96.0& 93.1           & 96.8	&93.2\\
     021\_bleach\_cleanser  & 96.0&  93.2& 96.8 &94.8& 96.8& \textbf{94.9}            &\textbf{97.1}&	94.0\\
     024\_bowl*             & 90.2 &  90.2& 96.3& 96.3& 96.9& 96.9           &\textbf{98.0}	&\textbf{98.0}\\
     025\_mug               & 97.6&  95.4& 97.3& 94.2& 97.6& 94.9         & \textbf{98.2}	&\textbf{95.7}\\
     035\_power\_drill      & 96.7&  95.1& 97.2& 95.9& 96.9& 95.2           & \textbf{98.3}&	\textbf{97.4}\\
     036\_wood\_block*      & 90.4&  90.4& 92.6& 92.6& 96.2& 96.2           & \textbf{97.0}	&\textbf{97.0}\\
     037\_scissors          & 96.7&  92.7& 97.7& 95.7& 97.2& 93.3            &\textbf{98.3}&	\textbf{96.8}\\
     040\_large\_marker     & 96.7&  91.8& 96.6& 89.1& 96.9 &92.7          & \textbf{98.6}	&\textbf{94.3}\\
     051\_large\_clamp*     & 93.6&  93.6&  \textbf{96.8}& \textbf{96.8}& 96.3 &96.3          & 95.9&	95.9\\
     052\_extra\_large\_clamp* & 88.4&  88.4& 96.0& 96.0 &\textbf{96.4}& \textbf{96.4}         & 95.6&	95.6\\
     061\_foam\_brick*          & 96.8&  96.8& 97.3& 97.3& 97.3 &97.3       & \textbf{98.6}	&\textbf{98.6}\\
     \hline
     mean & 95.5&  91.8&  96.6 &92.7& 96.7 &94.4             &\textbf{97.5} &\textbf{95.3} \\
    \bottomrule
  \end{tabular}
  \caption{\textbf{Comparison with State of the Art on YCB-V}. We report the Average Recall w.r.t AUC of ADD(-S) and AUC of ADD-S in \% and compare with state of the art. (*) denotes symmetric objects.}
  \label{tab:ycbv_full_results_AUC}
\end{table*}

\subsection{Qualitative Results}
\begin{figure*}
    \centering
    \includegraphics[width=1\linewidth]{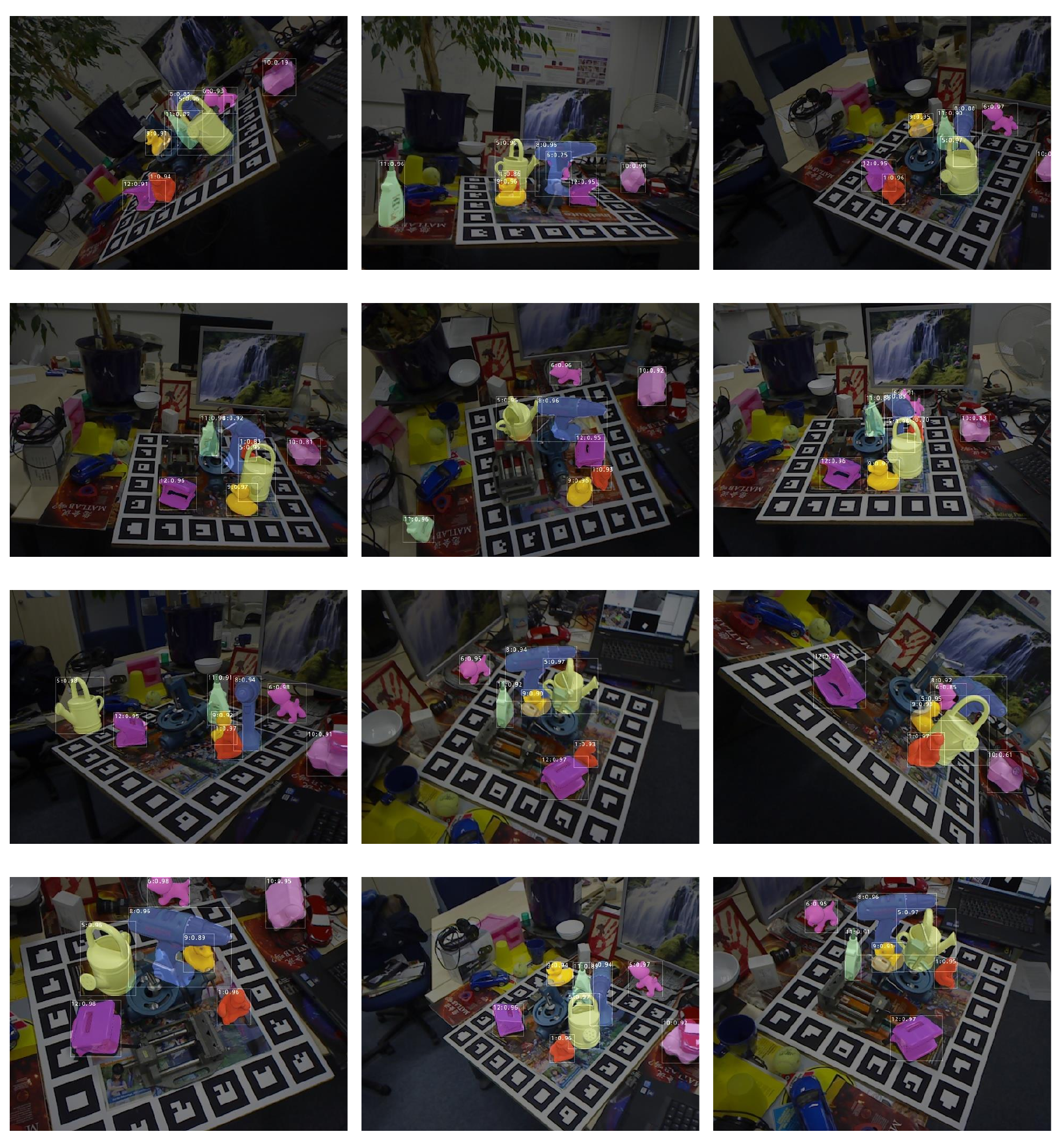}
    \caption{Qualitative Results on LM-O~\cite{Brachmann2016UncertaintyDriven6P}.}
    \label{fig:Qualitative_LMO}
\end{figure*}

\begin{figure*}
    \centering
    \includegraphics[width=1\linewidth]{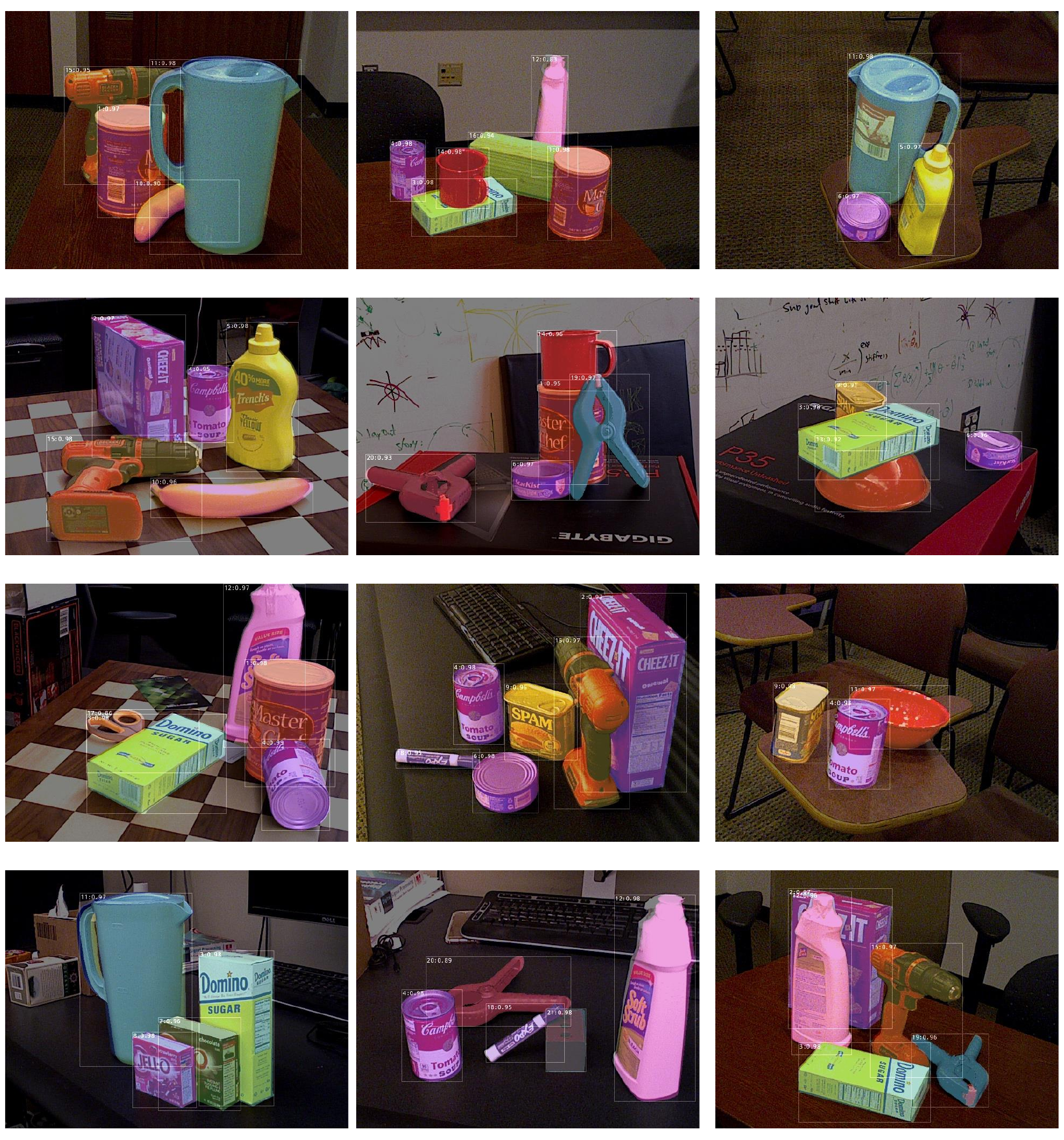}
    \caption{Qualitative Results on YCB-V~\cite{xiang2018posecnn}.}
    \label{fig:Qualitative_YCBV}
\end{figure*}

\begin{figure*}
    \centering
    \includegraphics[width=1\linewidth]{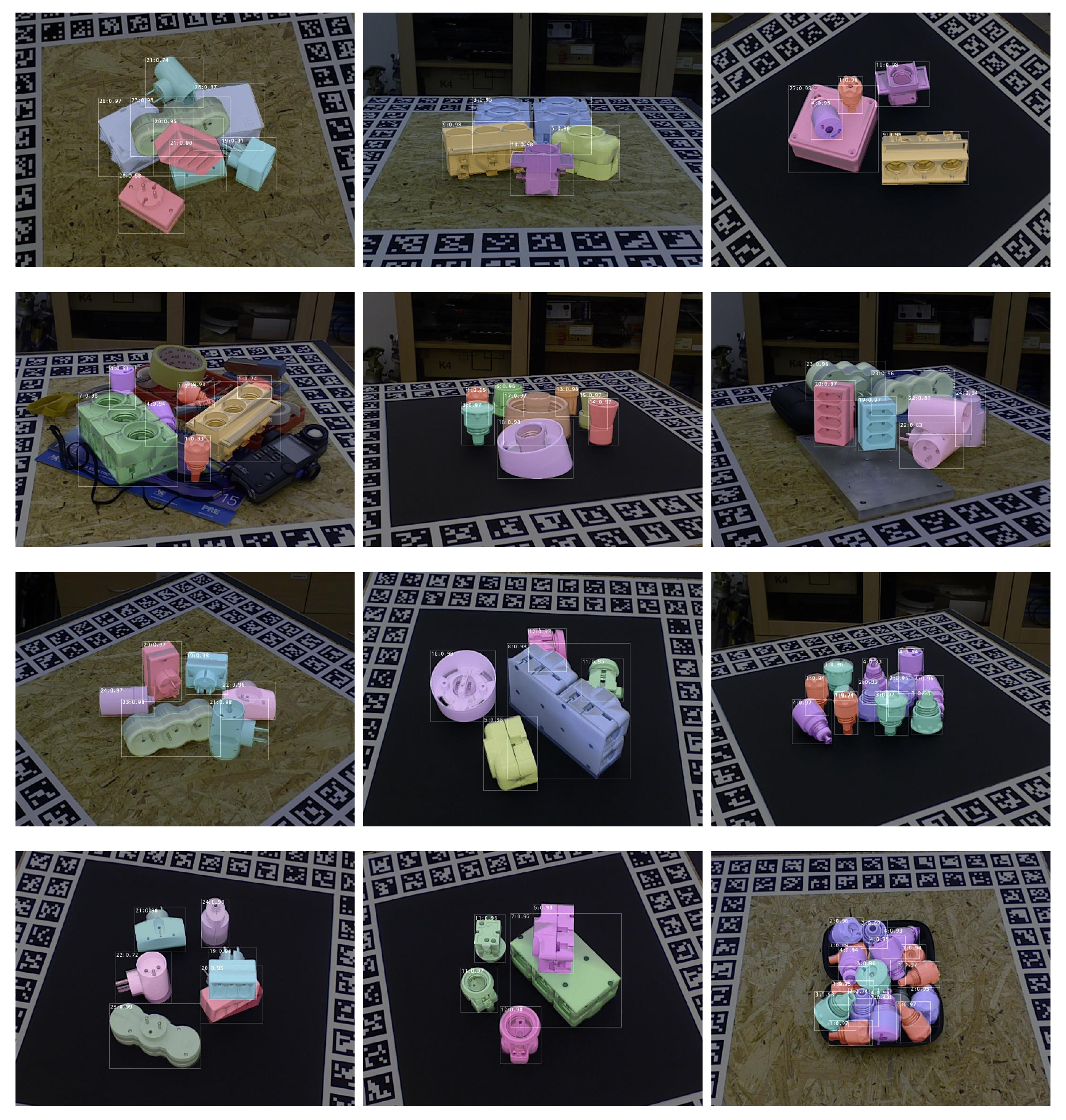}
    \caption{Qualitative Results on T-LESS~\cite{hodan2017t}.}
    \label{fig:Qualitative_TLESS}
\end{figure*}

We present quantitative results on the LM-O~\cite{Brachmann2016UncertaintyDriven6P}, YCB-V~\cite{xiang2018posecnn}, and T-LESS~\cite{hodan2017t} datasets in Figure~\ref{fig:Qualitative_LMO}, Figure~\ref{fig:Qualitative_YCBV}, and Figure~\ref{fig:Qualitative_TLESS}, respectively. We rendered the object into the image using the estimated pose. It is clear to see that the contour of the rendered object aligns seamlessly with the real object in the image, demonstrating the accuracy of our estimated pose. Furthermore, it is evident that our proposed HiPose performs well with texture-less objects and can handle occlusion effectively.

%% file: submission.bbl
\begin{thebibliography}{76}
\providecommand{\natexlab}[1]{#1}
\providecommand{\url}[1]{\texttt{#1}}
\expandafter\ifx\csname urlstyle\endcsname\relax
  \providecommand{\doi}[1]{doi: #1}\else
  \providecommand{\doi}{doi: \begingroup \urlstyle{rm}\Url}\fi

\bibitem[Aoki et~al.(2019)Aoki, Goforth, Srivatsan, and Lucey]{aoki2019pointnetlk}
Yasuhiro Aoki, Hunter Goforth, Rangaprasad~Arun Srivatsan, and Simon Lucey.
\newblock Pointnetlk: Robust \& efficient point cloud registration using pointnet.
\newblock In \emph{Proceedings of the IEEE/CVF conference on computer vision and pattern recognition}, pages 7163--7172, 2019.

\bibitem[Atzmon et~al.(2018)Atzmon, Maron, and Lipman]{atzmon2018point}
Matan Atzmon, Haggai Maron, and Yaron Lipman.
\newblock Point convolutional neural networks by extension operators.
\newblock \emph{arXiv preprint arXiv:1803.10091}, 2018.

\bibitem[Brachmann et~al.(2016)Brachmann, Michel, Krull, Yang, Gumhold, and Rother]{Brachmann2016UncertaintyDriven6P}
Eric Brachmann, Frank Michel, Alexander Krull, Michael~Ying Yang, Stefan Gumhold, and Carsten Rother.
\newblock Uncertainty-driven 6d pose estimation of objects and scenes from a single rgb image.
\newblock \emph{2016 IEEE Conference on Computer Vision and Pattern Recognition (CVPR)}, pages 3364--3372, 2016.

\bibitem[Cai et~al.(2022)Cai, Heikkil{\"a}, and Rahtu]{cai2022ove6d}
Dingding Cai, Janne Heikkil{\"a}, and Esa Rahtu.
\newblock Ove6d: Object viewpoint encoding for depth-based 6d object pose estimation.
\newblock In \emph{Proceedings of the IEEE/CVF Conference on Computer Vision and Pattern Recognition}, pages 6803--6813, 2022.

\bibitem[Chen et~al.(2020)Chen, Jia, Chang, Duan, and Leonardis]{chen2020g2l}
Wei Chen, Xi Jia, Hyung~Jin Chang, Jinming Duan, and Ales Leonardis.
\newblock G2l-net: Global to local network for real-time 6d pose estimation with embedding vector features.
\newblock In \emph{Proceedings of the IEEE/CVF conference on computer vision and pattern recognition}, pages 4233--4242, 2020.

\bibitem[Dang et~al.(2022)Dang, Wang, Guo, and Salzmann]{dang2022learning}
Zheng Dang, Lizhou Wang, Yu Guo, and Mathieu Salzmann.
\newblock Learning-based point cloud registration for 6d object pose estimation in the real world.
\newblock In \emph{European Conference on Computer Vision}, pages 19--37. Springer, 2022.

\bibitem[Di et~al.(2021)Di, Manhardt, Wang, Ji, Navab, and Tombari]{di2021so}
Yan Di, Fabian Manhardt, Gu Wang, Xiangyang Ji, Nassir Navab, and Federico Tombari.
\newblock So-pose: Exploiting self-occlusion for direct 6d pose estimation.
\newblock In \emph{Proceedings of the IEEE/CVF International Conference on Computer Vision}, pages 12396--12405, 2021.

\bibitem[Do et~al.(2018)Do, Cai, Pham, and Reid]{do2018deep6dpose}
Thanh-Toan Do, Ming Cai, Trung Pham, and Ian Reid.
\newblock Deep-6dpose: Recovering 6d object pose from a single rgb image.
\newblock \emph{arXiv preprint arXiv:1802.10367}, 2018.

\bibitem[Dosovitskiy et~al.(2020)Dosovitskiy, Beyer, Kolesnikov, Weissenborn, Zhai, Unterthiner, Dehghani, Minderer, Heigold, Gelly, Uszkoreit, and Houlsby]{Dosovitskiy2020AnII}
Alexey Dosovitskiy, Lucas Beyer, Alexander Kolesnikov, Dirk Weissenborn, Xiaohua Zhai, Thomas Unterthiner, Mostafa Dehghani, Matthias Minderer, Georg Heigold, Sylvain Gelly, Jakob Uszkoreit, and Neil Houlsby.
\newblock An image is worth 16x16 words: Transformers for image recognition at scale.
\newblock \emph{ArXiv}, abs/2010.11929, 2020.

\bibitem[Drost et~al.(2010)Drost, Ulrich, Navab, and Ilic]{drost2010model}
Bertram Drost, Markus Ulrich, Nassir Navab, and Slobodan Ilic.
\newblock Model globally, match locally: Efficient and robust 3d object recognition.
\newblock In \emph{2010 IEEE computer society conference on computer vision and pattern recognition}, pages 998--1005. Ieee, 2010.

\bibitem[Gao et~al.(2021)Gao, Lauri, Hu, Zhang, and Frintrop]{gao2021cloudaae}
Ge Gao, Mikko Lauri, Xiaolin Hu, Jianwei Zhang, and Simone Frintrop.
\newblock Cloudaae: Learning 6d object pose regression with on-line data synthesis on point clouds.
\newblock In \emph{2021 IEEE International Conference on Robotics and Automation (ICRA)}, pages 11081--11087. IEEE, 2021.

\bibitem[Ge et~al.(2021)Ge, Liu, Wang, Li, and Sun]{Ge2021YOLOXEY}
Zheng Ge, Songtao Liu, Feng Wang, Zeming Li, and Jian Sun.
\newblock Yolox: Exceeding yolo series in 2021.
\newblock \emph{ArXiv}, abs/2107.08430, 2021.

\bibitem[Hai et~al.(2023)Hai, Song, Li, Salzmann, and Hu]{Hai2023RigidityAwareDF}
Yang Hai, Rui Song, Jiaojiao Li, Mathieu Salzmann, and Yinlin Hu.
\newblock Rigidity-aware detection for 6d object pose estimation.
\newblock \emph{2023 IEEE/CVF Conference on Computer Vision and Pattern Recognition (CVPR)}, pages 8927--8936, 2023.

\bibitem[Haugaard and Buch(2021)]{Haugaard2021SurfEmbDA}
Rasmus~Laurvig Haugaard and Anders~Glent Buch.
\newblock Surfemb: Dense and continuous correspondence distributions for object pose estimation with learnt surface embeddings.
\newblock \emph{2022 IEEE/CVF Conference on Computer Vision and Pattern Recognition (CVPR)}, pages 6739--6748, 2021.

\bibitem[He et~al.(2015)He, Zhang, Ren, and Sun]{He2015DeepRL}
Kaiming He, X. Zhang, Shaoqing Ren, and Jian Sun.
\newblock Deep residual learning for image recognition.
\newblock \emph{2016 IEEE Conference on Computer Vision and Pattern Recognition (CVPR)}, pages 770--778, 2015.

\bibitem[He et~al.(2020)He, Sun, Huang, Liu, Fan, and Sun]{he2020pvn3d}
Yisheng He, Wei Sun, Haibin Huang, Jianran Liu, Haoqiang Fan, and Jian Sun.
\newblock Pvn3d: A deep point-wise 3d keypoints voting network for 6dof pose estimation.
\newblock In \emph{Proceedings of the IEEE/CVF conference on computer vision and pattern recognition}, pages 11632--11641, 2020.

\bibitem[He et~al.(2021)He, Huang, Fan, Chen, and Sun]{he2021ffb6d}
Yisheng He, Haibin Huang, Haoqiang Fan, Qifeng Chen, and Jian Sun.
\newblock Ffb6d: A full flow bidirectional fusion network for 6d pose estimation.
\newblock In \emph{Proceedings of the IEEE/CVF Conference on Computer Vision and Pattern Recognition}, pages 3003--3013, 2021.

\bibitem[Hodan et~al.(2017)Hodan, Haluza, Obdr{\v{z}}{\'a}lek, Matas, Lourakis, and Zabulis]{hodan2017t}
Tom{\'a}{\v{s}} Hodan, Pavel Haluza, {\v{S}}tep{\'a}n Obdr{\v{z}}{\'a}lek, Jiri Matas, Manolis Lourakis, and Xenophon Zabulis.
\newblock T-less: An rgb-d dataset for 6d pose estimation of texture-less objects.
\newblock In \emph{2017 IEEE Winter Conference on Applications of Computer Vision (WACV)}, pages 880--888. IEEE, 2017.

\bibitem[Hodan et~al.(2018)Hodan, Michel, Brachmann, Kehl, GlentBuch, Kraft, Drost, Vidal, Ihrke, Zabulis, et~al.]{hodan2018bop}
Tomas Hodan, Frank Michel, Eric Brachmann, Wadim Kehl, Anders GlentBuch, Dirk Kraft, Bertram Drost, Joel Vidal, Stephan Ihrke, Xenophon Zabulis, et~al.
\newblock Bop: Benchmark for 6d object pose estimation.
\newblock In \emph{Proceedings of the European conference on computer vision (ECCV)}, pages 19--34, 2018.

\bibitem[Hodan et~al.(2020)Hodan, Barath, and Matas]{hodan2020epos}
Tomas Hodan, Daniel Barath, and Jiri Matas.
\newblock Epos: Estimating 6d pose of objects with symmetries.
\newblock In \emph{Proceedings of the IEEE/CVF conference on computer vision and pattern recognition}, pages 11703--11712, 2020.

\bibitem[Hu et~al.(2020)Hu, Yang, Xie, Rosa, Guo, Wang, Trigoni, and Markham]{hu2020randla}
Qingyong Hu, Bo Yang, Linhai Xie, Stefano Rosa, Yulan Guo, Zhihua Wang, Niki Trigoni, and Andrew Markham.
\newblock Randla-net: Efficient semantic segmentation of large-scale point clouds.
\newblock In \emph{Proceedings of the IEEE/CVF conference on computer vision and pattern recognition}, pages 11108--11117, 2020.

\bibitem[Hu et~al.(2022)Hu, Fua, and Salzmann]{hu2022perspective}
Yinlin Hu, Pascal Fua, and Mathieu Salzmann.
\newblock Perspective flow aggregation for data-limited 6d object pose estimation.
\newblock In \emph{European Conference on Computer Vision}, pages 89--106. Springer, 2022.

\bibitem[Kehl et~al.(2017)Kehl, Manhardt, Tombari, Ilic, and Navab]{kehl2017ssd}
Wadim Kehl, Fabian Manhardt, Federico Tombari, Slobodan Ilic, and Nassir Navab.
\newblock Ssd-6d: Making rgb-based 3d detection and 6d pose estimation great again.
\newblock In \emph{Proceedings of the IEEE international conference on computer vision}, pages 1521--1529, 2017.

\bibitem[Kingma and Ba(2014)]{Kingma2014AdamAM}
Diederik~P. Kingma and Jimmy Ba.
\newblock Adam: A method for stochastic optimization.
\newblock \emph{CoRR}, abs/1412.6980, 2014.

\bibitem[K{\"o}nig and Drost(2020)]{konig2020hybrid}
Rebecca K{\"o}nig and Bertram Drost.
\newblock A hybrid approach for 6dof pose estimation.
\newblock In \emph{European Conference on Computer Vision}, pages 700--706. Springer, 2020.

\bibitem[Ku et~al.(2018)Ku, Mozifian, Lee, Harakeh, and Waslander]{ku2018joint}
Jason Ku, Melissa Mozifian, Jungwook Lee, Ali Harakeh, and Steven~L Waslander.
\newblock Joint 3d proposal generation and object detection from view aggregation.
\newblock In \emph{2018 IEEE/RSJ International Conference on Intelligent Robots and Systems (IROS)}, pages 1--8. IEEE, 2018.

\bibitem[Labb'e et~al.(2022)Labb'e, Manuelli, Mousavian, Tyree, Birchfield, Tremblay, Carpentier, Aubry, Fox, and Sivic]{Labbe2022MegaPose6P}
Yann Labb'e, Lucas Manuelli, Arsalan Mousavian, Stephen Tyree, Stan Birchfield, Jonathan Tremblay, Justin Carpentier, Mathieu Aubry, Dieter Fox, and Josef Sivic.
\newblock Megapose: 6d pose estimation of novel objects via render \& compare.
\newblock In \emph{Conference on Robot Learning}, 2022.

\bibitem[Lepetit et~al.(2009)Lepetit, Moreno-Noguer, and Fua]{lepetit2009ep}
Vincent Lepetit, Francesc Moreno-Noguer, and Pascal Fua.
\newblock Epnp: An accurate o (n) solution to the pnp problem.
\newblock \emph{International journal of computer vision}, 81:\penalty0 155--166, 2009.

\bibitem[Li et~al.(2018{\natexlab{a}})Li, Bai, and Hager]{li2018unified}
Chi Li, Jin Bai, and Gregory~D Hager.
\newblock A unified framework for multi-view multi-class object pose estimation.
\newblock In \emph{Proceedings of the european conference on computer vision (eccv)}, pages 254--269, 2018{\natexlab{a}}.

\bibitem[Li et~al.(2018{\natexlab{b}})Li, Bu, Sun, Wu, Di, and Chen]{li2018pointcnn}
Yangyan Li, Rui Bu, Mingchao Sun, Wei Wu, Xinhan Di, and Baoquan Chen.
\newblock Pointcnn: Convolution on x-transformed points.
\newblock \emph{Advances in neural information processing systems}, 31, 2018{\natexlab{b}}.

\bibitem[Li and Stamos(2023)]{li2023depth}
Zhujun Li and Ioannis Stamos.
\newblock Depth-based 6dof object pose estimation using swin transformer.
\newblock \emph{arXiv preprint arXiv:2303.02133}, 2023.

\bibitem[Li et~al.(2019)Li, Wang, and Ji]{li2019cdpn}
Zhigang Li, Gu Wang, and Xiangyang Ji.
\newblock Cdpn: Coordinates-based disentangled pose network for real-time rgb-based 6-dof object pose estimation.
\newblock In \emph{Proceedings of the IEEE/CVF International Conference on Computer Vision}, pages 7678--7687, 2019.

\bibitem[Liang et~al.(2018)Liang, Yang, Wang, and Urtasun]{liang2018deep}
Ming Liang, Bin Yang, Shenlong Wang, and Raquel Urtasun.
\newblock Deep continuous fusion for multi-sensor 3d object detection.
\newblock In \emph{Proceedings of the European conference on computer vision (ECCV)}, pages 641--656, 2018.

\bibitem[Lin et~al.(2023)Lin, Wang, Zhou, Liu, and Chen]{lin2023transpose}
Xiao Lin, Deming Wang, Guangliang Zhou, Chengju Liu, and Qijun Chen.
\newblock Transpose: 6d object pose estimation with geometry-aware transformer.
\newblock \emph{arXiv preprint arXiv:2310.16279}, 2023.

\bibitem[Lipson et~al.(2022)Lipson, Teed, Goyal, and Deng]{lipson2022coupled}
Lahav Lipson, Zachary Teed, Ankit Goyal, and Jia Deng.
\newblock Coupled iterative refinement for 6d multi-object pose estimation.
\newblock In \emph{Proceedings of the IEEE/CVF Conference on Computer Vision and Pattern Recognition}, pages 6728--6737, 2022.

\bibitem[Liu et~al.(2020)Liu, Jonschkowski, Angelova, and Konolige]{liu2020keypose}
Xingyu Liu, Rico Jonschkowski, Anelia Angelova, and Kurt Konolige.
\newblock Keypose: Multi-view 3d labeling and keypoint estimation for transparent objects.
\newblock In \emph{Proceedings of the IEEE/CVF conference on computer vision and pattern recognition}, pages 11602--11610, 2020.

\bibitem[Liu et~al.(2022{\natexlab{a}})Liu, Zhang, Zhang, Fu, Tang, Liang, Tang, Cheng, Zhang, Wang, and Ji]{liu2022gdrnpp_bop}
Xingyu Liu, Ruida Zhang, Chenyangguang Zhang, Bowen Fu, Jiwen Tang, Xiquan Liang, Jingyi Tang, Xiaotian Cheng, Yukang Zhang, Gu Wang, and Xiangyang Ji.
\newblock Gdrnpp.
\newblock \url{https://github.com/shanice-l/gdrnpp_bop2022}, 2022{\natexlab{a}}.

\bibitem[Liu et~al.(2022{\natexlab{b}})Liu, Mao, Wu, Feichtenhofer, Darrell, and Xie]{Liu2022ACF}
Zhuang Liu, Hanzi Mao, Chaozheng Wu, Christoph Feichtenhofer, Trevor Darrell, and Saining Xie.
\newblock A convnet for the 2020s.
\newblock \emph{2022 IEEE/CVF Conference on Computer Vision and Pattern Recognition (CVPR)}, pages 11966--11976, 2022{\natexlab{b}}.

\bibitem[Manhardt et~al.(2019)Manhardt, Kehl, and Gaidon]{manhardt2019roi}
Fabian Manhardt, Wadim Kehl, and Adrien Gaidon.
\newblock Roi-10d: Monocular lifting of 2d detection to 6d pose and metric shape.
\newblock In \emph{Proceedings of the IEEE/CVF Conference on Computer Vision and Pattern Recognition}, pages 2069--2078, 2019.

\bibitem[Oberweger et~al.(2018)Oberweger, Rad, and Lepetit]{oberweger2018making}
Markus Oberweger, Mahdi Rad, and Vincent Lepetit.
\newblock Making deep heatmaps robust to partial occlusions for 3d object pose estimation.
\newblock In \emph{Proceedings of the European Conference on Computer Vision (ECCV)}, pages 119--134, 2018.

\bibitem[Park et~al.(2019)Park, Patten, and Vincze]{park2019pix2pose}
Kiru Park, Timothy Patten, and Markus Vincze.
\newblock Pix2pose: Pixel-wise coordinate regression of objects for 6d pose estimation.
\newblock In \emph{Proceedings of the IEEE/CVF International Conference on Computer Vision}, pages 7668--7677, 2019.

\bibitem[Peng et~al.(2019)Peng, Liu, Huang, Zhou, and Bao]{peng2019pvnet}
Sida Peng, Yuan Liu, Qixing Huang, Xiaowei Zhou, and Hujun Bao.
\newblock Pvnet: Pixel-wise voting network for 6dof pose estimation.
\newblock In \emph{Proceedings of the IEEE/CVF Conference on Computer Vision and Pattern Recognition}, pages 4561--4570, 2019.

\bibitem[Qi et~al.(2017)Qi, Yi, Su, and Guibas]{qi2017pointnet++}
Charles~Ruizhongtai Qi, Li Yi, Hao Su, and Leonidas~J Guibas.
\newblock Pointnet++: Deep hierarchical feature learning on point sets in a metric space.
\newblock \emph{Advances in neural information processing systems}, 30, 2017.

\bibitem[Qi et~al.(2018)Qi, Liu, Wu, Su, and Guibas]{qi2018frustum}
Charles~R Qi, Wei Liu, Chenxia Wu, Hao Su, and Leonidas~J Guibas.
\newblock Frustum pointnets for 3d object detection from rgb-d data.
\newblock In \emph{Proceedings of the IEEE conference on computer vision and pattern recognition}, pages 918--927, 2018.

\bibitem[Rambach et~al.(2017)Rambach, Pagani, and Stricker]{rambach2017poster}
Jason Rambach, Alain Pagani, and Didier Stricker.
\newblock [poster] augmented things: Enhancing ar applications leveraging the internet of things and universal 3d object tracking.
\newblock In \emph{2017 IEEE International Symposium on Mixed and Augmented Reality (ISMAR-Adjunct)}, pages 103--108. IEEE, 2017.

\bibitem[Rothganger et~al.(2006)Rothganger, Lazebnik, Schmid, and Ponce]{rothganger20063d}
Fred Rothganger, Svetlana Lazebnik, Cordelia Schmid, and Jean Ponce.
\newblock 3d object modeling and recognition using local affine-invariant image descriptors and multi-view spatial constraints.
\newblock \emph{International journal of computer vision}, 66:\penalty0 231--259, 2006.

\bibitem[Rusinkiewicz and Levoy(2001)]{Rusinkiewicz2001EfficientVO}
Szymon Rusinkiewicz and Marc Levoy.
\newblock Efficient variants of the icp algorithm.
\newblock \emph{Proceedings Third International Conference on 3-D Digital Imaging and Modeling}, pages 145--152, 2001.

\bibitem[Su et~al.(2019)Su, Rambach, Minaskan, Lesur, Pagani, and Stricker]{su2019deep}
Yongzhi Su, Jason Rambach, Nareg Minaskan, Paul Lesur, Alain Pagani, and Didier Stricker.
\newblock Deep multi-state object pose estimation for augmented reality assembly.
\newblock In \emph{2019 IEEE International Symposium on Mixed and Augmented Reality Adjunct (ISMAR-Adjunct)}, pages 222--227. IEEE, 2019.

\bibitem[Su et~al.(2021)Su, Rambach, Pagani, and Stricker]{su2021synpo}
Yongzhi Su, Jason Rambach, Alain Pagani, and Didier Stricker.
\newblock Synpo-net—accurate and fast cnn-based 6dof object pose estimation using synthetic training.
\newblock \emph{Sensors}, 21\penalty0 (1):\penalty0 300, 2021.

\bibitem[Su et~al.(2022)Su, Saleh, Fetzer, Rambach, Navab, Busam, Stricker, and Tombari]{su2022zebrapose}
Yongzhi Su, Mahdi Saleh, Torben Fetzer, Jason Rambach, Nassir Navab, Benjamin Busam, Didier Stricker, and Federico Tombari.
\newblock Zebrapose: Coarse to fine surface encoding for 6dof object pose estimation.
\newblock In \emph{Proceedings of the IEEE/CVF Conference on Computer Vision and Pattern Recognition}, pages 6738--6748, 2022.

\bibitem[Su et~al.(2023)Su, Di, Zhai, Manhardt, Rambach, Busam, Stricker, and Tombari]{su2023opa}
Yongzhi Su, Yan Di, Guangyao Zhai, Fabian Manhardt, Jason Rambach, Benjamin Busam, Didier Stricker, and Federico Tombari.
\newblock Opa-3d: Occlusion-aware pixel-wise aggregation for monocular 3d object detection.
\newblock \emph{IEEE Robotics and Automation Letters}, 8\penalty0 (3):\penalty0 1327--1334, 2023.

\bibitem[Sundermeyer et~al.(2018)Sundermeyer, Marton, Durner, Brucker, and Triebel]{sundermeyer2018implicit}
Martin Sundermeyer, Zoltan-Csaba Marton, Maximilian Durner, Manuel Brucker, and Rudolph Triebel.
\newblock Implicit 3d orientation learning for 6d object detection from rgb images.
\newblock In \emph{Proceedings of the european conference on computer vision (ECCV)}, pages 699--715, 2018.

\bibitem[Sundermeyer et~al.(2020)Sundermeyer, Durner, Puang, Marton, Vaskevicius, Arras, and Triebel]{sundermeyer2020multi}
Martin Sundermeyer, Maximilian Durner, En~Yen Puang, Zoltan-Csaba Marton, Narunas Vaskevicius, Kai~O Arras, and Rudolph Triebel.
\newblock Multi-path learning for object pose estimation across domains.
\newblock In \emph{Proceedings of the IEEE/CVF conference on computer vision and pattern recognition}, pages 13916--13925, 2020.

\bibitem[Sundermeyer et~al.(2023)Sundermeyer, Hodan, Labb{\'e}, Wang, Brachmann, Drost, Rother, and Matas]{Sundermeyer2023BOPC2}
Martin Sundermeyer, Tom{\'a}s Hodan, Yann Labb{\'e}, Gu Wang, Eric Brachmann, Bertram Drost, Carsten Rother, and Juan E.~Sala Matas.
\newblock Bop challenge 2022 on detection, segmentation and pose estimation of specific rigid objects.
\newblock \emph{2023 IEEE/CVF Conference on Computer Vision and Pattern Recognition Workshops (CVPRW)}, pages 2785--2794, 2023.

\bibitem[Tian et~al.(2019)Tian, Shen, Chen, and He]{Tian2019FCOSFC}
Zhi Tian, Chunhua Shen, Hao Chen, and Tong He.
\newblock Fcos: Fully convolutional one-stage object detection.
\newblock \emph{2019 IEEE/CVF International Conference on Computer Vision (ICCV)}, pages 9626--9635, 2019.

\bibitem[Umeyama(1991)]{umeyama1991least}
Shinji Umeyama.
\newblock Least-squares estimation of transformation parameters between two point patterns.
\newblock \emph{IEEE Transactions on Pattern Analysis \& Machine Intelligence}, 13\penalty0 (04):\penalty0 376--380, 1991.

\bibitem[Wada et~al.(2020)Wada, Sucar, James, Lenton, and Davison]{wada2020morefusion}
Kentaro Wada, Edgar Sucar, Stephen James, Daniel Lenton, and Andrew~J Davison.
\newblock Morefusion: Multi-object reasoning for 6d pose estimation from volumetric fusion.
\newblock In \emph{Proceedings of the IEEE/CVF conference on computer vision and pattern recognition}, pages 14540--14549, 2020.

\bibitem[Wang et~al.(2019)Wang, Xu, Zhu, Mart{\'\i}n-Mart{\'\i}n, Lu, Fei-Fei, and Savarese]{wang2019densefusion}
Chen Wang, Danfei Xu, Yuke Zhu, Roberto Mart{\'\i}n-Mart{\'\i}n, Cewu Lu, Li Fei-Fei, and Silvio Savarese.
\newblock Densefusion: 6d object pose estimation by iterative dense fusion.
\newblock In \emph{Proceedings of the IEEE/CVF conference on computer vision and pattern recognition}, pages 3343--3352, 2019.

\bibitem[Wang et~al.(2021)Wang, Manhardt, Tombari, and Ji]{wang2021gdr}
Gu Wang, Fabian Manhardt, Federico Tombari, and Xiangyang Ji.
\newblock Gdr-net: Geometry-guided direct regression network for monocular 6d object pose estimation.
\newblock In \emph{Proceedings of the IEEE/CVF Conference on Computer Vision and Pattern Recognition}, pages 16611--16621, 2021.

\bibitem[Wang and Solomon(2019{\natexlab{a}})]{wang2019deep}
Yue Wang and Justin~M Solomon.
\newblock Deep closest point: Learning representations for point cloud registration.
\newblock In \emph{Proceedings of the IEEE/CVF international conference on computer vision}, pages 3523--3532, 2019{\natexlab{a}}.

\bibitem[Wang and Solomon(2019{\natexlab{b}})]{wang2019prnet}
Yue Wang and Justin~M Solomon.
\newblock Prnet: Self-supervised learning for partial-to-partial registration.
\newblock \emph{Advances in neural information processing systems}, 32, 2019{\natexlab{b}}.

\bibitem[Wu et~al.(2022)Wu, Zand, Etemad, and Greenspan]{wu2022vote}
Yangzheng Wu, Mohsen Zand, Ali Etemad, and Michael Greenspan.
\newblock Vote from the center: 6 dof pose estimation in rgb-d images by radial keypoint voting.
\newblock In \emph{European Conference on Computer Vision}, pages 335--352. Springer, 2022.

\bibitem[Xiang et~al.(2018)Xiang, Schmidt, Narayanan, and Fox]{xiang2018posecnn}
Yu Xiang, Tanner Schmidt, Venkatraman Narayanan, and Dieter Fox.
\newblock Posecnn: A convolutional neural network for 6d object pose estimation in cluttered scenes.
\newblock 2018.

\bibitem[Xu et~al.(2018)Xu, Anguelov, and Jain]{xu2018pointfusion}
Danfei Xu, Dragomir Anguelov, and Ashesh Jain.
\newblock Pointfusion: Deep sensor fusion for 3d bounding box estimation.
\newblock In \emph{Proceedings of the IEEE conference on computer vision and pattern recognition}, pages 244--253, 2018.

\bibitem[Yew and Lee(2020)]{yew2020rpm}
Zi~Jian Yew and Gim~Hee Lee.
\newblock Rpm-net: Robust point matching using learned features.
\newblock In \emph{Proceedings of the IEEE/CVF conference on computer vision and pattern recognition}, pages 11824--11833, 2020.

\bibitem[Yuan et~al.(2020)Yuan, Eckart, Kim, Jampani, Fox, and Kautz]{yuan2020deepgmr}
Wentao Yuan, Benjamin Eckart, Kihwan Kim, Varun Jampani, Dieter Fox, and Jan Kautz.
\newblock Deepgmr: Learning latent gaussian mixture models for registration.
\newblock In \emph{Computer Vision--ECCV 2020: 16th European Conference, Glasgow, UK, August 23--28, 2020, Proceedings, Part V 16}, pages 733--750. Springer, 2020.

\bibitem[Zakharov et~al.(2019)Zakharov, Shugurov, and Ilic]{zakharov2019dpod}
Sergey Zakharov, Ivan Shugurov, and Slobodan Ilic.
\newblock Dpod: 6d pose object detector and refiner.
\newblock In \emph{Proceedings of the IEEE/CVF international conference on computer vision}, pages 1941--1950, 2019.

\bibitem[Zhai et~al.(2023{\natexlab{a}})Zhai, Cai, Huang, Di, Manhardt, Tombari, Navab, and Busam]{zhai2023sg}
Guangyao Zhai, Xiaoni Cai, Dianye Huang, Yan Di, Fabian Manhardt, Federico Tombari, Nassir Navab, and Benjamin Busam.
\newblock Sg-bot: Object rearrangement via coarse-to-fine robotic imagination on scene graphs.
\newblock \emph{arXiv preprint arXiv:2309.12188}, 2023{\natexlab{a}}.

\bibitem[Zhai et~al.(2023{\natexlab{b}})Zhai, Huang, Wu, Jung, Di, Manhardt, Tombari, Navab, and Busam]{zhai2023monograspnet}
Guangyao Zhai, Dianye Huang, Shun-Cheng Wu, HyunJun Jung, Yan Di, Fabian Manhardt, Federico Tombari, Nassir Navab, and Benjamin Busam.
\newblock Monograspnet: 6-dof grasping with a single rgb image.
\newblock In \emph{2023 IEEE International Conference on Robotics and Automation (ICRA)}, pages 1708--1714. IEEE, 2023{\natexlab{b}}.

\bibitem[Zhang et~al.(2023)Zhang, Di, Zhang, Zhai, Manhardt, Tombari, and Ji]{zhang2023ddf}
Chenyangguang Zhang, Yan Di, Ruida Zhang, Guangyao Zhai, Fabian Manhardt, Federico Tombari, and Xiangyang Ji.
\newblock Ddf-ho: Hand-held object reconstruction via conditional directed distance field.
\newblock \emph{arXiv preprint arXiv:2308.08231}, 2023.

\bibitem[Zhao et~al.(2017)Zhao, Shi, Qi, Wang, and Jia]{zhao2017pyramid}
Hengshuang Zhao, Jianping Shi, Xiaojuan Qi, Xiaogang Wang, and Jiaya Jia.
\newblock Pyramid scene parsing network.
\newblock In \emph{Proceedings of the IEEE conference on computer vision and pattern recognition}, pages 2881--2890, 2017.

\bibitem[Zhao et~al.(2020)Zhao, Zhang, Guan, Zhao, Peng, and Fan]{zhao2020learning}
Wanqing Zhao, Shaobo Zhang, Ziyu Guan, Wei Zhao, Jinye Peng, and Jianping Fan.
\newblock Learning deep network for detecting 3d object keypoints and 6d poses.
\newblock In \emph{Proceedings of the IEEE/CVF Conference on computer vision and pattern recognition}, pages 14134--14142, 2020.

\bibitem[Zhou et~al.(2020)Zhou, Yan, Wang, and Chen]{zhou2020novel}
Guangliang Zhou, Yi Yan, Deming Wang, and Qijun Chen.
\newblock A novel depth and color feature fusion framework for 6d object pose estimation.
\newblock \emph{IEEE Transactions on Multimedia}, 23:\penalty0 1630--1639, 2020.

\bibitem[Zhou et~al.(2021)Zhou, Wang, Chen, and Huang]{Zhou2021PRGCNAD}
Guangyuan Zhou, Huiqun Wang, Jiaxin Chen, and Di Huang.
\newblock Pr-gcn: A deep graph convolutional network with point refinement for 6d pose estimation.
\newblock \emph{2021 IEEE/CVF International Conference on Computer Vision (ICCV)}, pages 2773--2782, 2021.

\bibitem[Zhou et~al.(2023)Zhou, Chen, Xu, Dou, and Qin]{zhou2023deep}
Jun Zhou, Kai Chen, Linlin Xu, Qi Dou, and Jing Qin.
\newblock Deep fusion transformer network with weighted vector-wise keypoints voting for robust 6d object pose estimation.
\newblock In \emph{Proceedings of the IEEE/CVF International Conference on Computer Vision}, pages 13967--13977, 2023.

\bibitem[Zhou et~al.(2018)Zhou, Park, and Koltun]{zhou2018open3d}
Qian-Yi Zhou, Jaesik Park, and Vladlen Koltun.
\newblock Open3d: A modern library for 3d data processing.
\newblock \emph{arXiv preprint arXiv:1801.09847}, 2018.

\end{thebibliography}
